\documentclass{article}

 \PassOptionsToPackage{numbers, compress}{natbib}

\usepackage[preprint]{neurips_2024}

\usepackage[utf8]{inputenc} 
\usepackage[T1]{fontenc}    
\usepackage{hyperref}       
\usepackage{url}            
\usepackage{booktabs}       
\usepackage{amsfonts}       
\usepackage{nicefrac}       
\usepackage{microtype}      
\usepackage{xcolor}         

\usepackage{amsmath}
\usepackage{amssymb}
\usepackage{booktabs}

\usepackage[accsupp]{axessibility}  
\usepackage{algorithm}
\usepackage{amsfonts,amssymb} 
\usepackage{amsmath}
\usepackage{color}
\usepackage{multirow}
\usepackage{pythonhighlight}
\usepackage{setspace}
\usepackage{amssymb}
\usepackage{multicol}
\usepackage{amssymb}
\usepackage{bbm}
\usepackage{makecell}
\usepackage{pifont}

\usepackage{url}            
\usepackage{booktabs}       
\usepackage{amsfonts}       
\usepackage{nicefrac}       
\usepackage{microtype}      
\usepackage{xcolor}         
\usepackage{amsmath}
\usepackage[pdftex]{graphicx}

\usepackage{multirow}
\usepackage{colortbl}
\usepackage{enumitem}
\usepackage{wrapfig}
\usepackage{algorithm}
\usepackage{algpseudocode}
\usepackage[figuresright]{rotating}
\usepackage{makecell}
\usepackage{hyperref}

\newcommand{\ie}{{\it i.e.}}
\newcommand{\eg}{{\it e.g.}}

\newcommand{\bd}[4]{\textbf{#1#2#3#4}}

\definecolor{grey1}{RGB}{230, 230, 230}
\definecolor{my_yellow}{RGB}{204, 167, 68}
\definecolor{my_white}{RGB}{193, 141, 158}

\title{Small Object Few-shot Segmentation for Vision-based Industrial Inspection}

\author{%
  Zilong Zhang$^1$,$\quad$ Chang Niu$^2$, $\quad$ Zhibin Zhao$^1$, $\quad$ Xingwu Zhang$^1$\thanks{Corresponding author.}, $\quad$ Xuefeng Chen$^1$ \\
  $^1$ National Key Lab of Aerospace Power System and Plasma Technology, Xi’an Jiaotong University \\
  $^2$ School of Electronic and Information Engineering, South China University of Technology \\
  \texttt{zhangzilongc@gmail.com,eeniu@mail.scut.edu.cn}, \\
  \texttt{\{zhaozhibin,xwzhang,chenxf\}@xjtu.edu.cn} \\
}

\begin{document}

\maketitle

\begin{abstract}
Vision-based industrial inspection (VII) aims to locate defects quickly and accurately. Supervised learning under a close-set setting and industrial anomaly detection, as two common paradigms in VII, face different problems in practical applications. The former is that various and sufficient defects are difficult to obtain, while the latter is that specific defects cannot be located. To solve these problems, in this paper, we focus on the few-shot semantic segmentation (FSS) method, which can locate unseen defects conditioned on a few annotations without retraining. Compared to common objects in natural images, the defects in VII are small. This brings two problems to current FSS methods: \ding{172} distortion of target semantics and \ding{173} many false positives for backgrounds. To alleviate these problems, we propose a \textbf{s}mall \textbf{o}bject \textbf{f}ew-shot \textbf{s}egmentation (SOFS) model. The key idea for alleviating \ding{172} is to avoid the resizing of the original image and correctly indicate the intensity of target semantics. SOFS achieves this idea via the non-resizing procedure and the prototype intensity downsampling of support annotations. To alleviate \ding{173}, we design an abnormal prior map in SOFS to guide the model to reduce false positives and propose a mixed normal Dice loss to preferentially prevent the model from predicting false positives. SOFS can achieve FSS and few-shot anomaly detection determined by support masks. Diverse experiments substantiate the superior performance of SOFS. Code is available at \href{https://github.com/zhangzilongc/SOFS}{https://github.com/zhangzilongc/SOFS}.
\end{abstract}

\section{Introduction}
\label{1_introduction}

\begin{figure}[t]
	\centering
	\includegraphics[width=\linewidth]{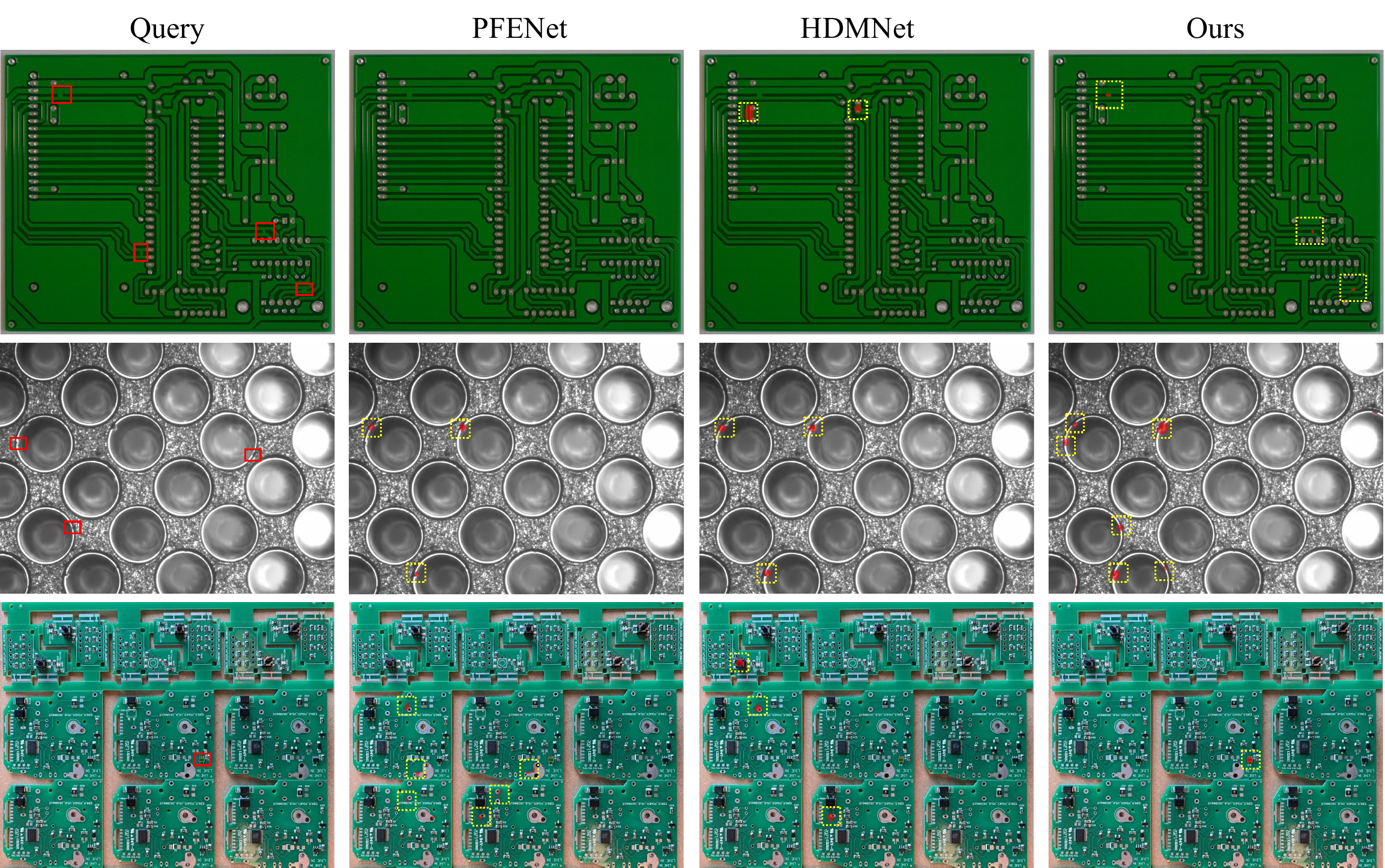}
	\caption{Small defects predictions of different models conditioned on the same support image. The red solid line box indicates true labels, and the yellow dashed line box indicates predictions. Due to the small defects, please use the electronic version to enlarge defects for easier viewing.}
	\label{fig_1}	
\end{figure}

Vision-based industrial inspection (VII) automatically identifies defects in industrial processes and ensures product quality. Compared to object recognition in natural images, defect recognition in VII is fine-grained. As shown in Fig.\ref{fig_1}, fine-grained defects appear as small objects, the area proportions of defects are mostly less than 0.3\%, and the smallest is even 0.015\%. 

The common paradigm in VII is supervised learning under a close-set setting \cite{qu2021deeply, gao2022cas, zou2018deepcrack, liu2021crackformer, zeng2022small}. This paradigm requires researchers to collect a large number of various defects in specific scenarios, and then design specific models for fitting the distribution of small defects in training. Since industrial processes are generally optimized to minimize the production of defective products \cite{yang2020dfr}, collecting various and sufficient defects for training is often intractable, which makes it difficult for such methods to be widely applied. To tackle this problem, industrial anomaly detection (IAD) \cite{salehi2021multiresolution, deng2022anomaly, roth2022towards, tien2023revisiting, zhang2023exploring, you2022unified} has been widely adopted recently. IAD only requires normal samples to detect defects, while there are a large number of normal samples. Such a characteristic is conducive to the application of IAD in real scenarios. However, due to limitations of the problem formulation of IAD, it regards various defects as the same class, \ie, abnormal class, which prevents it from locating specific defects. In real industrial processes, different defects have significantly different impacts on products, \eg, scratches on the printed circuit board affect the appearance, while broken circuits affect the use. Ablation of aircraft engine blades affects engine efficiency, and cracks may cause engine fire \cite{zhang2023industrial}. Thus, locating specific defects has significant practical value. 

To break through the above limitations, this paper focuses on the few-shot semantic segmentation (FSS) method. FSS divides the input into the query and support sets following the episode paradigm \cite{vinyals2016matching}. It segments the query targets conditioned on the semantic clues from a few support annotations and can quickly adapt to unseen classes without retraining. These properties make FSS an ideal approach for VII. The achievement of previous FSS methods \cite{lang2023base, tian2020prior, yang2020prototype, peng2023hierarchical, sun2024vrp, xu2023self} is built upon the accurate extraction of semantic features of the support set, determined by the feature intensity in the downsampling of support annotations and the information in the resizing image. As shown in Fig. \ref{fig_1}, applying these methods to segment small defects brings two problems: failure to segment and many false positives for backgrounds. The former is due to the distortion of target semantics, caused by the information loss during the resizing of the original image and the feature intensity bias in the downsampling of support annotations. The latter is caused by the missing defect, \eg, the first and third rows in Fig. \ref{fig_1}, and the train-set overfitting.


In this paper, to alleviate the above problems, we propose a \textbf{s}mall \textbf{o}bject \textbf{f}ew-shot \textbf{s}egmentation (SOFS) model for VII. SOFS alleviates the distortion of small objects by 1) avoiding the resizing and 2) correctly indicating the intensity of target semantics. Specifically, 1) SOFS uses a non-resizing procedure to crop small objects in training and adopt the sliding window mechanism in the test. SOFS ensures that the pixel area of small objects encoded by the model is consistent with that in the original image. 2) We propose the prototype intensity downsampling of support annotations to correct the bias caused by the common interpolation. This downsampling is calculated by all pixels of corresponding regions of target semantics, providing a better estimate of target semantics.

To reduce false positives caused by the missing defect and the train-set overfitting, 1) We design an abnormal prior map in SOFS to guide the model to highlight the semantic while ignoring the normal background. This design also enables SOFS to have both FSS and few-shot anomaly detection (FAD) abilities. 2) We propose a mixed normal Dice loss to impose a large penalty when the model predicts false positives, preferentially preventing the model from predicting false positives.

Contributions: 1) We recognize FSS as an ideal solution for addressing the difficulty of obtaining diverse defects and the inability of anomaly detection to identify specific defects in VII. 2) We propose SOFS to alleviate the distortion of target semantics and many false positives. The key idea of the former is to avoid the resizing of the original image and correctly indicate the intensity of target semantics. The latter is to guide the model to focus more on backgrounds. SOFS can adapt to unseen classes without retraining, achieving FSS and FAD. To the best of our knowledge, SOFS is the first model to realize both FSS and FAD in VII. 3) FSS experiments on VISION V1 show that SOFS achieves state-of-the-art results, achieving $12.2\%$ mean mIoU improvements compared to SegGPT. FAD experiments also demonstrate its competitive performance.

\section{Related Work}
\label{2_related}

\subsection{Few-shot Semantic Segmentation}

Few-shot semantic segmentation \cite{zhang2022feature, cheng2022holistic, lang2023base, zhang2019canet} predicts dense masks for novel classes with only a few annotations. According to the different fusion granularity of support features and query features, previous methods can be divided into image-level feature fusion \cite{lang2023base, zhang2019canet, tian2020prior, wang2019panet, yang2020prototype} or pixel-level feature fusion \cite{peng2023hierarchical, sun2024vrp, xu2023self}. The image-level feature fusion methods extract the semantic clue of support sets by a masked average pooling of a global image and then fuse this prototype with the query features. Since a pooling prototype feature cannot cover all regions of an object, the pixel-level feature fusion methods are proposed to mine the correspondence between the query pixel-level features and the support semantic-related pixel-level features, where the residual connection in the cross attention plays the role of fusing query and support features.


Previous methods mostly focus on the segmentation of PASCAL-$5^{i}$ \cite{shaban2017one, everingham2010pascal} and COCO-$20^{i}$ \cite{nguyen2019feature, lin2014microsoft}, where the segmented objects are mostly salient. In comparison, the segmented objects in VII are generally small. In VISION V1 \cite{bai2023vision}, $84\%$ of defect types have area proportions less than $0.3\%$, and $38\%$ of defect types have pixel areas less than $900$. These characteristics either cause previous methods to predict many false positives or fail to segment.


\subsection{Small Object Recognition}

The term “small object” refers to objects that occupy a small fraction of the input image \cite{rekavandi2023transformers}, \eg, in the widely used MS COCO dataset \cite{lin2014microsoft}, it defines objects whose bounding box is $32\times 32$ pixels or less in a typical $480 \times 640$ image, while other datasets define the objects that occupy 10\% of the image. The main challenges for small object recognition include information loss, low tolerance for bounding box perturbation, etc \cite{cheng2023towards}. Information loss refers to the fact that the feature information of small objects is almost wiped out during the downsampling of the feature extractor, it has the greatest impact on performance. To alleviate this issue, there are mainly three kinds of methods. The first is to maintain the high-resolution feature map while retaining the fast process \cite{xu2022fea, yang2019clustered, duan2021coarse}. The second is a super-resolution-based method \cite{haris2021task, bai2018sod, li2017perceptual}, which restores the distorted structures of small objects. The third is to process small objects by multi-scale learning and hierarchical feature fusion \cite{tan2020efficientdet, zhao2019m2det}.

Previous methods mainly focus on small object recognition in a close-set setting. In this paper, we study this in an open-set setting, the target is to learn the meta-class-agnostic feature.

\begin{figure}[t]
	\centering
	\includegraphics[width=\linewidth]{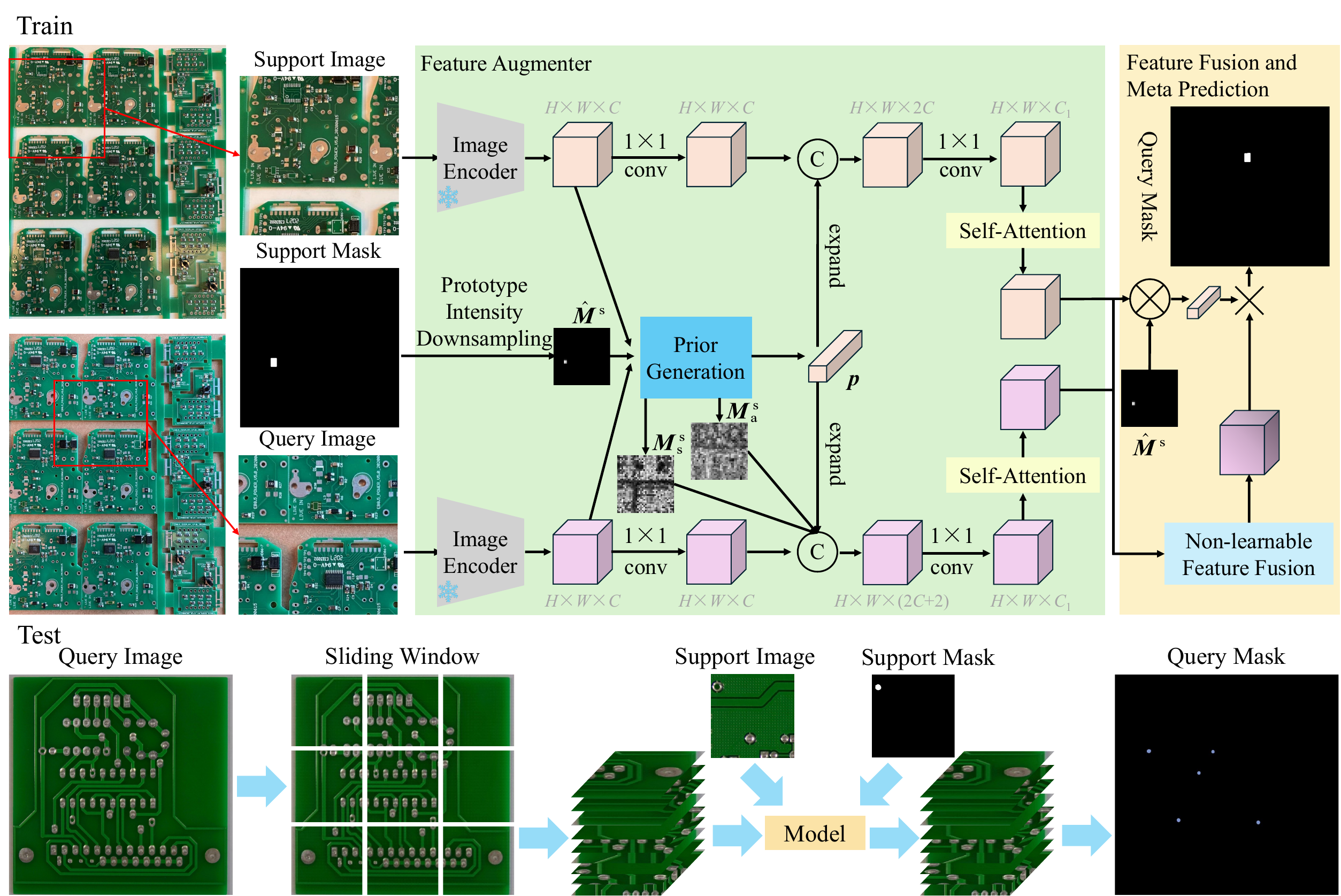}
	\caption{Proposed small object few-shot segmentation model. $\hat{\textbf{\textit{M}}}^{\rm{s}}$ indicates the downsampling of support mask. $\textbf{\textit{M}}^{\rm{s}}_{\rm{a}}, \textbf{\textit{M}}^{\rm{s}}_{\rm{s}}$ denote the abnormal prior map and semantic prior map respectively, $\textbf{\textit{p}}$ denotes a prototype feature. Best viewed in the electronic version.}
	\label{fig_2}	
\end{figure}


\subsection{Few-shot Anomaly Detection (Abnormal Segmentation)}

Recently, few-shot anomaly detection (FAD) \cite{fang2023fastrecon, huang2022registration, santos2023optimizing, zhu2024toward, jeong2023winclip} has been proposed to simultaneously solve the problem of difficulty in collecting various and sufficient defects in VII and the problem of requiring retraining for different objects. FAD can quickly adapt to unknown classes by a few normal samples, \ie, support images. FAD methods detect anomalies based on the similarity of pixel-level features between the query image and the support images. Furthermore, recent works \cite{jeong2023winclip, li2023myriad, chen2023zero} add textual descriptions to supplement the missing details of the image modality.

Anomaly detection regards various defects as the same class, \ie, abnormal class, which prevents it from identifying specific defects. In VII, different defects have significantly different impacts on products, which makes anomaly detection unsuitable. In this paper, our proposed method can simultaneously achieve FSS and FAD and is applicable in more industrial scenarios.

\section{Methods}
\label{3_method}

\subsection{Problem Formulation}

Few-shot semantic segmentation trains model to capture the meta-knowledge of the training set, adapting to novel objects with only a few annotated support images. In definition, the model is trained on the training set $D_{train}$ and tested on the test set $D_{test}$. Suppose the category sets in $D_{train}$ and $D_{test}$ are $C_{train}$ and $C_{test}$ respectively. In the few-shot setting, $C_{train}\cap C_{test}=\emptyset$. Episodes are applied to $D_{train}$ and $D_{test}$, each episode consists of a query set $\textbf{\textit{Q}}=\{ {(\textbf{\textit{I}}^{\rm{q}}, \textbf{\textit{M}}^{\rm{q}})}\}$ and a support set $\textbf{\textit{S}}=\{{(\textbf{\textit{I}}^{\rm{s}}_{i}, \textbf{\textit{M}}^{\rm{s}}_{i})}\}^{K}_{i=1}$ with the same class, where $\textbf{\textit{I}}^{\rm{q}}, \textbf{\textit{I}}^{\rm{s}}_{i} \in \mathbb{R}^{\bar{H}\times \bar{W}\times 3}$ represent the RGB images and $\textbf{\textit{M}}^{\rm{q}}, \textbf{\textit{M}}^{\rm{s}}_{i} \in \{0, 1 \}^{\bar{H}\times \bar{W}}$ denote their binary masks, $K$ denotes the number of support images. In training, both $\textbf{\textit{M}}^{\rm{q}}$ and $\textbf{\textit{M}}^{\rm{s}}_{i}$ are used, while only $\textbf{\textit{M}}^{\rm{s}}_{i}$ are accessible in testing. During testing, the model requires no optimization for the novel classes. In FAD, $\textbf{\textit{M}}^{\rm{s}}_{i} \in \{0 \}^{\bar{H}\times \bar{W}}$.


\subsection{Small Object Few-shot Segmentation}
\subsubsection{Overview}

The proposed small object few-shot segmentation (SOFS) model consists of three major steps: 1) small object enhancement designs, 2) feature augmenter, 3) feature fusion and meta prediction, as illustrated in Fig. \ref{fig_2}, where the small object enhancement designs alleviate the distortion of small objects. In training, SOFS crops the semantic area on the support images and query images (Section \ref{ED}). The cropped query/support images are then encoded by a feature augmenter to generate the semantic-related features (Section \ref{FA}). Afterward, the encoded query and support features are fused via a non-learnable feature fusion. Finally, the fused feature interacts with the prototype features to produce the query mask (Section \ref{FFMP}). In testing, SOFS uses a sliding window mechanism to produce multiple query images. To reduce false positives caused by the train-set overfitting, we elaborate on a mixed normal Dice loss (Section \ref{LF}).


\subsubsection{Small Object Enhancement Designs}
\label{ED}


\textbf{Non-resizing Procedure}. The core idea is to ensure that the pixel area of small objects encoded by the model is consistent with that in the original image. As shown in Fig. \ref{fig_2}, the non-resizing procedure randomly crops the small object on the original image in training and uses the sliding window mechanism to process all regions of the query image in the test. The reason why the non-resizing procedure can be used is that the segmentation of small defects in VII has a strong correlation with the surrounding areas, but a weak correlation with distant areas, which eliminates the need for the model to establish long-range dependencies. In addition to the above benefits, the non-resizing procedure can produce normal samples (the non-defective areas are all normal areas). We show in Section \ref{LF} how these normal samples alleviate model overfitting. 

In the following part, we refer to query set and support set after the non-resizing procedure as $\textbf{\textit{Q}}=\{ {(\textbf{\textit{I}}^{\rm{q}}, \textbf{\textit{M}}^{\rm{q}})}\}$ and $\textbf{\textit{S}}=\{ {(\textbf{\textit{I}}^{\rm{s}}_{(1)}, \textbf{\textit{M}}^{\rm{s}}_{(1)})}\}$, where we show the case $K=1$ without loss of generality.

\begin{figure}[t]
	\centering
	\includegraphics[width=\linewidth]{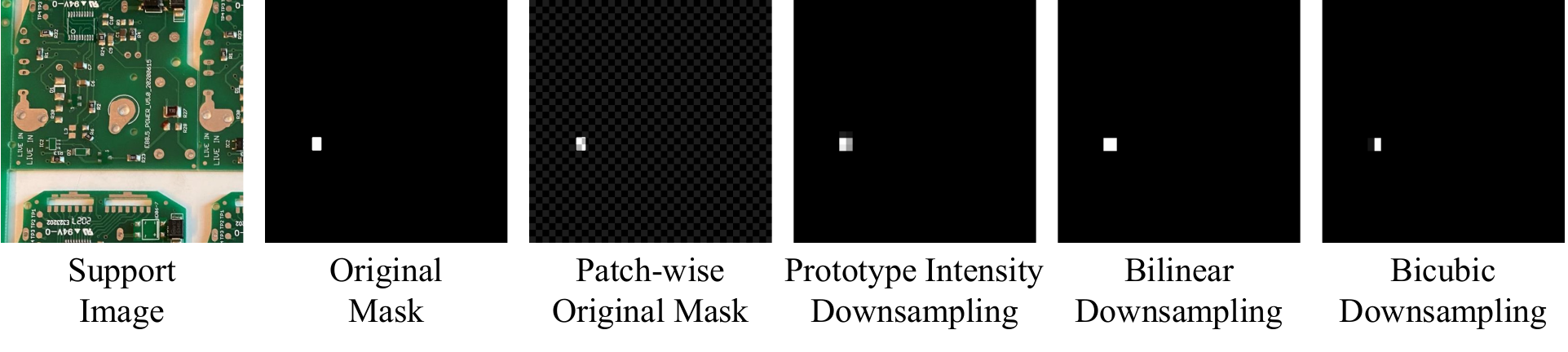}
	\caption{Different downsamplings for the small object mask. Each patch on the patch-wise original mask represents a region that is highly correlated with corresponding features on the feature map.}
	\label{fig_3}
\end{figure}

\textbf{Prototype Intensity Downsampling}. To extract the prototype feature on the support image, we need to downsample the support mask $\textbf{\textit{M}}^{\rm{s}}$ to ensure that it is consistent with the size of the support feature map. As shown in the patch-wise original mask of Fig. \ref{fig_3}, every patch represents a region that is highly correlated with corresponding features on the feature map, since the segmented object is small, some corresponding features of the feature map may be weak, occupying only a small part of the original segmentation area. However, if we use the common bilinear/bicubic interpolation to downsample the original mask to indicate the feature intensity, since the bilinear/bicubic interpolation only uses $4$/$16$ points to get the result, the result may be overestimated or underestimated, as shown in Fig. \ref{fig_3}, which leads to the distortion of target semantics. To alleviate this issue, we propose a prototype intensity downsampling to replace the common bilinear/bicubic interpolation in the downsampling of $\textbf{\textit{M}}^{\rm{s}}$. Specifically, we employ a $l\times l$ convolution layer to process $\textbf{\textit{M}}^{\rm{s}}$: 

\begin{equation}
\label{EN:1}
\hat{\textbf{\textit{M}}}^{\rm{s}}=\frac{Conv(\textbf{\textit{M}}^{\rm{s}})}{l\times l} ,
\end{equation}

where $\hat{\textbf{\textit{M}}}^{\rm{s}}\in \mathbb{R}^{H\times W}$ denotes the downsampling of the support mask, $l$ denotes the multiple of downsampling of feature map, the parameters in the convolution kernel are all 1, the stride of convolution is $l$. $\hat{\textbf{\textit{M}}}^{\rm{s}}$ is calculated by all pixels of corresponding regions of features. It provides a better estimate of the intensity of prototype features, avoiding the mismatch of semantic clues.

\subsubsection{Feature Augmenter}
\label{FA}

Firstly, we encode $\textbf{\textit{I}}^{\rm{s}}$ and $\textbf{\textit{I}}^{\rm{q}}$ separately, resulting in feature maps $\textbf{\textit{F}}^{\rm{s}}, \textbf{\textit{F}}^{\rm{q}}\in \mathbb{R}^{H\times W\times C}$. Subsequently, we use a prior generation module to generate the prototype feature $\textbf{\textit{p}}\in \mathbb{R}^{C}$, a semantic prior map $\textbf{\textit{M}}^{\rm{q}}_{\rm{s}}\in \mathbb{R}^{H\times W}$ and an abnormal prior map $\textbf{\textit{M}}^{\rm{q}}_{\rm{a}}\in \mathbb{R}^{H\times W}$. $\textbf{\textit{p}}$ is extracted by a $MaskAvgPool(\textbf{\textit{F}}^{\rm{s}},\hat{\textbf{\textit{M}}}^{\rm{s}})$, where $MaskAvgPool(\cdot)$ denotes the masked average pooling (detailed descriptions are provided in Appendix). $\textbf{\textit{M}}^{\rm{q}}_{\rm{s}}$ and $\textbf{\textit{M}}^{\rm{q}}_{\rm{a}}$ are formulated as follows:


\begin{gather}
\textbf{\textit{M}}^{\rm{q}}_{\rm{s}}=\Psi(\Phi(\textbf{\textit{F}}^{\rm{q}}) \Phi(\textbf{\textit{F}}^{\rm{s}})^{\rm{T}}\odot \varphi(\hat{\textbf{\textit{M}}}^{\rm{s}})), \label{Eq.4} 
\\
\textbf{\textit{M}}^{\rm{q}}_{\rm{a}}=1-\Psi(\Phi(\textbf{\textit{F}}^{\rm{q}}) \Phi(\textbf{\textit{F}}^{\rm{s}})^{\rm{T}}\odot (1-\varphi(\hat{\textbf{\textit{M}}}^{\rm{s}}))),  \label{Eq.5} 
\end{gather}

where $\Phi(\cdot): \mathbb{R}^{H\times W\times C}\mapsto \mathbb{R}^{HW\times C}$ refers to the reshape function and $\rm{L}_{2}$ normalized on feature dimension, $\varphi(\cdot): \mathbb{R}^{H\times W}\mapsto \mathbb{R}^{1\times HW}\mapsto \mathbb{R}^{HW\times HW}$ refers to the reshape and repeat along the row, $\Psi(\cdot): \mathbb{R}^{HW\times HW}\mapsto \mathbb{R}^{HW\times 1} \mapsto \mathbb{R}^{H\times W}$ refers to taking the maximum value along the column and the reshape. $\textbf{\textit{M}}^{\rm{q}}_{\rm{s}}$ denotes the maximum similarity between each pixel-level query feature and support semantic features. The larger the value on $\textbf{\textit{M}}^{\rm{q}}_{\rm{s}}$, the more relevant it is to the semantics in the support set. $1-\textbf{\textit{M}}^{\rm{q}}_{\rm{a}}$ denotes the maximum similarity between each pixel-level query feature and support normal features. The larger the value on $\textbf{\textit{M}}^{\rm{q}}_{\rm{a}}$, the smaller the similarity with normal features in the support set, and the more abnormal the corresponding region is.


The motivation for adding $\textbf{\textit{M}}^{\rm{q}}_{\rm{a}}$ is that we hope that SOFS can reduce false positives for backgrounds in VII. Specifically, in VII, there is a unique type of defect: the missing defect, where the target area misses some components, \eg, the defect shown in Fig. \ref{fig_2} is a missing defect, we can only observe the image background. If we only use semantic matching for the missing defect, it may highlight many false backgrounds. $\textbf{\textit{M}}^{\rm{q}}_{\rm{a}}$ matches every pixel-level query feature with the normal support features, if there is a missing defect, it can be highlighted and the normal background can not be. In addition, $\textbf{\textit{M}}^{\rm{q}}_{\rm{a}}$ enables SOFS to have FAD ability, we can input the normal support image.

After obtaining the above features, we concat $\textbf{\textit{F}}^{\rm{s}}$ with $\textbf{\textit{p}}$, concat $\textbf{\textit{F}}^{\rm{q}}$ with $\textbf{\textit{p}}, \textbf{\textit{M}}^{\rm{q}}_{\rm{a}}$, and $\textbf{\textit{M}}^{\rm{q}}_{\rm{s}}$, then use a $1\times 1$ convolution to produce $\bar{\textbf{\textit{F}}}^{\rm{s}}, \bar{\textbf{\textit{F}}}^{\rm{q}}\in \mathbb{R}^{H\times W\times C_{1}}$. Subsequently, we use the self-attention \cite{vaswani2017attention} to further improve $\bar{\textbf{\textit{F}}}^{\rm{s}}, \bar{\textbf{\textit{F}}}^{\rm{q}}$.

\subsubsection{Feature Fusion and Meta Prediction}
\label{FFMP}

$\bar{\textbf{\textit{F}}}^{\rm{s}}, \bar{\textbf{\textit{F}}}^{\rm{q}}$ are subsequently fused by a non-learnable feature fusion. The matching mechanism follows \cite{peng2023hierarchical} to replace the dot produce with the cosine similarities, which is formulated as follows:

\begin{equation}
\label{EN:5}
\bar{\textbf{\textit{F}}}^{\rm{qs}}=\rm{softmax}^{\prime}(\frac{\Phi(\bar{\textbf{\textit{F}}}^{\rm{q}}) \Phi(\bar{\textbf{\textit{F}}}^{\rm{s}})^{\rm{T}}}{\tau}){\Phi}^{\prime}(\bar{\textbf{\textit{F}}}^{\rm{s}}\odot \vartheta(\hat{\textbf{\textit{M}}}^{\rm{s}}))+{\Phi}^{\prime}(\bar{\textbf{\textit{F}}}^{\rm{q}}),
\end{equation}

where ${\Phi}^{\prime}(\cdot): \mathbb{R}^{H\times W\times C}\mapsto \mathbb{R}^{HW\times C}$ refers to the reshape function and non $\rm{L}_{2}$ normalized, $\tau$ controls the distribution shape, $\rm{softmax}^{\prime}(\cdot)$ refers to the normalization along the row, \ie, reverse softmax \cite{smith2017offline}, $\vartheta(\cdot): \mathbb{R}^{H\times W} \mapsto \mathbb{R}^{H\times W\times C}$ refers to first expanding the new dimension and then replicating along the expanded dimension. Eq. (\ref{EN:5}) can be regarded as a type of cross-attention, where the learnable parameters are discarded. We think that the recognition of small objects does not need lots of parameters, more parameters may cause the risk of overfitting the category-specific information. After the feature fusion, we extract the prototype feature $\bar{\textbf{\textit{p}}}\in \mathbb{R}^{C_{1}}$ by a $MaskAvgPool(\bar{\textbf{\textit{F}}}^{\rm{s}}, \hat{\textbf{\textit{M}}}^{\rm{s}})$. Then the prediction result of SOFS model is formulated as follows:

\begin{equation}
\label{EN:6}
\hat{\textbf{\textit{M}}}^{\rm{q}}_{\rm{pred}}=\mathbb{I}(\rm{sum}(\textbf{\textit{M}}^{\rm{s}})>0) \rm{Sigmoid}(\phi(\bar{\textbf{\textit{F}}}^{\rm{qs}}\bar{\textbf{\textit{p}}}))+(1-\mathbb{I}(\rm{sum}(\textbf{\textit{M}}^{\rm{s}})>0))\textbf{\textit{M}}^{\rm{q}}_{\rm{a}},
\end{equation}

where $\mathbb{I}(\cdot)$ refers to a indicator function, $\rm{Sigmoid}(\cdot)$ is a sigmoid function, $\phi(\cdot): \mathbb{R}^{HW\times 1}\mapsto \mathbb{R}^{H\times W}$ refers to the reshape function. Compared with the learnable prediction classifier (a multi-layer perceptron) used in \cite{peng2023hierarchical, tian2020prior}, our meta prediction can be learned dynamically, it is encouraged to learn a class-agnostic meta feature. $\hat{\textbf{\textit{M}}}^{\rm{q}}_{\rm{pred}}$ is a combination of a semantics prediction and an abnormal prior map, if the support images are all normal samples, SOFS produces a result of few-shot abnormal segmentation. Otherwise, SOFS produces a result of few-shot semantic segmentation.

\subsubsection{Loss Function}
\label{LF}

In our experiments, we find that training SOFS by Dice loss \cite{milletari2016v} and the cross entropy loss suffers from a severe overfitting problem, producing many false positives. To alleviate this issue, we propose a mixed normal Dice loss as follows:

\begin{equation}
\label{EN:7}
\rm{DiceLoss}_{normal}(\textbf{\textit{X}}, \textbf{\textit{Y}})= \textit{t}(1-\frac{2 |\textbf{\textit{X}} \cap \textbf{\textit{Y}}|}{|\textbf{\textit{X}}| + |\textbf{\textit{Y}} |}) + \beta(1-\textit{t})( 1- \frac{1}{\eta|\textbf{\textit{X}} | + 1}),
\end{equation}

where $\textit{t}=\mathbb{I}(\rm{sum}(\textbf{\textit{Y}})>0)$, $\textbf{\textit{X}}, \textbf{\textit{Y}}$ refers to the prediction and the ground truth, $\eta$ is a large hyperparameter, \eg, $1\rm{e}^{4}$, $\beta$ is a hyperparameter to control the weight, we set $\beta=1$ in the experiment. The normal sample term (the second term of Eq. (\ref{EN:7})) can be regarded as a regularization term of the original Dice loss. At the beginning of training, the model almost outputs 0 for all inputs. As the training evolves, once the model has a false prediction for the normal sample, the backpropagation of this term produces a larger gradient compared to defective samples, which preferentially prevents the model from predicting false positives. This loss facilitates that the model does not produce false positives while predicting the defects as much as possible. In our implementation, the normal query samples are from the non-defective areas in the non-resizing procedure, the probability of sampling normal query samples is $\alpha$.

Our final loss function is a linear combination of the cross entropy loss and the mixed normal Dice loss, where the weight of the cross entropy loss is $\gamma$.

\section{Experiments}
\label{4_exp}

\subsection{Implementation Details}

\textbf{Datasets.} \textbf{FSS:} We conduct experiments on VISION V1 \cite{bai2023vision} and Defect Spectrum \cite{yang2023defect}. VISION V1 \cite{bai2023vision} released in CVPR 2023 Vision-based InduStrial InspectiON workshop \footnote{\href{https://vision-based-industrial-inspection.github.io/cvpr-2023/}{https://vision-based-industrial-inspection.github.io/cvpr-2023/}}, is one of the largest and most diverse manufacturing datasets in terms of both image number and available annotations. VISION V1 consists of 14 objects, spanning 44 defect types. $84\%$ of defect types have area proportions less than $0.3\%$, and $38\%$ of defect types have pixel areas less than $900$.  Defect Spectrum \cite{yang2023defect} relabeled the data from MVTec AD \cite{bergmann2019mvtec}, DAGM2007 \cite{wieler2007weakly}, part of VISION V1 \cite{bai2023vision} and Cotton-Fabric \cite{fabric}, including 97 defect types. Defect Spectrum is the dataset with the largest number of defect types so far. $63\%$ of defect types have area proportions less than $2\%$. For the above datasets, cross-validation is conducted by dividing all objects into 3 folds, we ensure that the distributions of area proportions of defects in each fold are roughly the same. In the experiments, two folds serve as training data, while the remaining one is used for testing. To ensure performance stability and fairness for comparison, the query/support pairs sampling by all methods are the same. \textbf{FAD:}We use two benchmark FAD datasets, \ie, MVTec AD \cite{bergmann2019mvtec} and VisA \cite{zou2022spot}, including 27 objects in total. More details of datasets and experiments are shown in Appendix.

\textbf{Metrics.} \textbf{FSS:} We adopt the mean intersection over union (mIoU) as the evaluation metric. We denote $\rm{mIoU}=1/C\Sigma^{C}_{i=1}\rm{IoU}_{i}$, where $C$ is the number of defect classes in each fold, and $\rm{IoU}_{i}$ indicates intersection-over-union for class $i$. \textbf{FAD:} For classification, we report Area Under the Receiver Operating Characteristic (AUROC). For segmentation, we report pixel-wise AUROC and pixel-wise F$_{1}$-max.

\textbf{Training and Test Details.} \textbf{FSS:} We use ResNet-50 \cite{he2016deep} and DINO v2 Base \cite{oquab2023dinov2} as the encoder to verify the effectiveness of the proposed method and reimplementation methods on different backbones. The training input sizes of ResNet-50 and DINO v2 Base are set to $512\times 512$ and $518\times 518$ respectively. $l$ in the prototype intensity downsampling are set to $8$ and $14$ for ResNet-50 and DINO v2 Base respectively. The probability $\alpha$ is set to $0.3$, $\eta$ is set to $1\rm{e}^{5}$, $\gamma$ is set to $0.01$, $\tau$ is set to 0.1. In the non-resizing procedure, the stride of sliding window and the input size in test are the same as the input size in training. All reimplementation methods and SOFS use the same training augmentations. \textbf{FAD:} We use DINO v2 Base \cite{oquab2023dinov2} as the encoder. The input size is set to $518\times 518$. We only test SOFS. More details of SOFS and reimplementation methods are shown in Appendix.

\begin{table}[!t]
  \caption{Few-shot semantic segmentation performance on VISION V1 \cite{bai2023vision}. We report 1-shot and 5-shot results using the mIoU (\%). “NR”: non-resizing procedure. “MNDL”: mixed normal Dice loss. “F0”: Fold 0. The results are the average of 15 runs.}
  \label{tab:vision}
  \centering
  \small
  \setlength\tabcolsep{5.4pt}%
  \begin{tabular}{clcccccccccc}
    \toprule
    \multirow{2}{*}{Backbone} & \multirow{2}{*}{Method} & \multicolumn{4}{c}{1-shot} & \multicolumn{4}{c}{5-shot} \\
    \cmidrule{3-10}
      &  & F0 & F1 & F2 & Mean & F0 & F1 & F2 & Mean \\
      \midrule
    \multirow{5}{*}{ResNet 50} & PFENet~\cite{tian2020prior} {\fontsize{5pt}{2pt}\selectfont TPAMI' 2020} & 4.1 & 2.5 & 6.1 & 4.2 & 5.1 & 2.9 & 8.1& 5.4 \\
    & PFENet+NR+MNDL & 6.2 & 3.1 & 7.0 & 5.4 & 7.0 & 3.5 & 7.5 & 6.0 \\
    & HDMNet~\cite{peng2023hierarchical} {\fontsize{5pt}{2pt}\selectfont CVPR' 2023} & 4.4 & 4.0 & 8.2 & 5.5 & 5.9 & 8.8 & 11.0 & 8.6 \\
    & HDMNet+NR+MNDL & 9.3 & 4.5 & 6.2 & 6.7 & 11.1 & 9.8 & 8.5 & 9.8 \\
    & \cellcolor{grey1} SOFS~(Ours) & \cellcolor{grey1} 10.3 & \cellcolor{grey1} 4.5 & \cellcolor{grey1}10.2 & \cellcolor{grey1} 8.3 & \cellcolor{grey1} 14.9 & \cellcolor{grey1} 5.9 & \cellcolor{grey1} 15.5 & \cellcolor{grey1} 12.1 \\
    \midrule
    \multirow{5}{*}{DINO v2 Base} & PFENet~\cite{tian2020prior} {\fontsize{5pt}{2pt}\selectfont TPAMI' 2020} & 6.7 & 3.5 & 13.2 & 7.8 & 7.3 & 4.6 & 13.8 & 8.6 \\
    & PFENet+NR+MNDL & 8.5 & 6.9 & 9.8 & 8.4 & 9.6 & 7.1 & 11.4 & 9.4 \\
    & HDMNet~\cite{peng2023hierarchical} {\fontsize{5pt}{2pt}\selectfont CVPR' 2023} & 8.2 & 5.9 & 10.7 & 8.3 & 10.6 & 11.0 & 15.9 & 12.5 \\
    & HDMNet+NR+MNDL & 18.5 & 15.3 & 12.0 & 15.3 & 22.9 & 18.9 & 14.9 & 18.9 \\
    & \cellcolor{grey1} SOFS~(Ours) & \cellcolor{grey1} \bd22.9 & \cellcolor{grey1} \bd23.1 & \cellcolor{grey1} \bd21.6 & \cellcolor{grey1} \bd22.5 & \cellcolor{grey1} \bd29.2 & \cellcolor{grey1} \bd27.0 & \cellcolor{grey1} 20.6 & \cellcolor{grey1} \bd25.6 \\
    \midrule
    Training Free & \\
    \multirow{2}{*}{ViT Large} & SegGPT~\cite{wang2023seggpt} {\fontsize{5pt}{2pt}\selectfont ICCV' 2023} & 2.9 & 10.5 & 20.0 & 11.1 & 3.6 & 13.0 & 23.8 & 13.4 \\
    & SegGPT+NR & 8.0 & 11.8 & 19.0 & 12.9 & 9.3 & 17.1 & \bd24.0 & 16.8 \\
    \bottomrule
  \end{tabular}
\end{table}

\begin{table}[!t]
  \caption{Few-shot semantic segmentation performance on Defect Spectrum \cite{yang2023defect}. We report 1-shot and 5-shot results using the mIoU (\%). The results are the average of 15 runs.}
  \label{tab:ds_spec}
  \centering
  \small
  \setlength\tabcolsep{5.4pt}%
  \begin{tabular}{clcccccccccc}
    \toprule
    \multirow{2}{*}{Backbone} & \multirow{2}{*}{Method} & \multicolumn{4}{c}{1-shot} & \multicolumn{4}{c}{5-shot} \\
    \cmidrule{3-10}
      &  & F0 & F1 & F2 & Mean & F0 & F1 & F2 & Mean \\
      \midrule
    \multirow{5}{*}{ResNet 50} & PFENet~\cite{tian2020prior} {\fontsize{5pt}{2pt}\selectfont TPAMI' 2020} & 9.5 & 14.6 & 16.7 & 13.6 & 10.5 & 18.0 & 19.2 & 15.9 \\
    & PFENet+NR+MNDL & 10.6 & 15.6 & 16.5 & 14.2 & 11.4 & 18.7 & 17.3 & 15.8 \\
    & HDMNet~\cite{peng2023hierarchical} {\fontsize{5pt}{2pt}\selectfont CVPR' 2023} & 13.5 & 22.0 & 22.6 & 19.4 & 18.4 & 28.3 & 27.9 & 24.9 \\
    & HDMNet+NR+MNDL & 14.8 & 18.1 & 23.4 & 18.8 & 18.7 & 25.8 & 26.6 & 23.7 \\
    & \cellcolor{grey1} SOFS~(Ours) & \cellcolor{grey1} 19.8 & \cellcolor{grey1} 22.8 & \cellcolor{grey1} 28.0 & \cellcolor{grey1} 23.5 & \cellcolor{grey1} 26.8 & \cellcolor{grey1} 27.7 & \cellcolor{grey1} 32.6 & \cellcolor{grey1} 29.0 \\
    \midrule
    \multirow{5}{*}{DINO v2 Base} & PFENet~\cite{tian2020prior} {\fontsize{5pt}{2pt}\selectfont TPAMI' 2020} & 27.5 & 33.0 & 37.1 & 32.5 & 29.7 & 34.9 & 39.4 & 34.7 \\
    & PFENet+NR+MNDL & 28.2 & 34.7 & 36.4 & 33.1 & 29.7 & 34.4 & 38.1 & 34.1 \\
    & HDMNet~\cite{peng2023hierarchical} {\fontsize{5pt}{2pt}\selectfont CVPR' 2023} & 30.1 & 34.5 & 39.7 & 34.8 & 34.8 & 35.9 & \bd 40.9 & 37.2 \\
    & HDMNet+NR+MNDL & 32.2 & 36.8 & 38.7 & 35.9 & 36.9 & 41.1 & 40.2 & 39.4 \\
    & \cellcolor{grey1} SOFS~(Ours) & \cellcolor{grey1} \bd35.6 & \cellcolor{grey1} \bd41.8 & \cellcolor{grey1} \bd39.9 & \cellcolor{grey1} \bd39.1 & \cellcolor{grey1} \bd40.0 & \cellcolor{grey1} 42.9 & \cellcolor{grey1} 40.0 & \cellcolor{grey1} \bd41.0 \\
    \midrule
    Training Free & \\
    \multirow{2}{*}{ViT Large} & SegGPT~\cite{wang2023seggpt} {\fontsize{5pt}{2pt}\selectfont ICCV' 2023} & 27.7 & 36.7 & 30.0 & 31.5 & 32.0 & \bd44.0 & 34.7 & 36.9 \\
    & SegGPT+NR & 30.1 & 34.8 & 28.5 & 31.1 & 33.6 & 40.6 & 32.6 & 35.6 \\
    \bottomrule
  \end{tabular}
\end{table}

\subsection{Comparison with State-of-the-Art Methods}

\textbf{FSS:} In Table \ref{tab:vision}, we report the comparison of SOFS with other state-of-the-art FSS approaches on VISION V1. Compared with other methods, SOFS can gain passable improvement on ResNet-50. However, due to the insufficient discriminability of pre-trained features for small defects, the performance is extremely low. As we choose DINO v2 Base \cite{oquab2023dinov2} as the encoder, the performance has a significant improvement and outperforms the others by a large margin. Despite using a model with fewer parameters, SOFS significantly outperforms SegGPT, achieving $11.4\%$ (1-shot) and $12.2\%$ (5-shot) of mean mIoU improvements. Notably, when applying the non-resizing procedure and the mixed normal Dice loss to comparison methods, the performance improves consistently, this illustrates the generality of the proposed method. In addition, we show qualitative results of SOFS in Fig. \ref{fig_1}, \ref{fig_6}, \ref{fig_7}. Although there are different defect types in one image, SOFS can segment the specific small defects. Table \ref{tab:ds_spec} shows the FSS performance on Defect Spectrum. Although this dataset contains larger defects than VISION V1, SOFS still achieves the best in most cases. The limitation of SOFS is that if the support images increase (K increases), the performance improvement for large defects is relatively weak (VISION V1 F2, DS-Spectrum F1 and F2). We think this is due to the underfitting to large defects. \textbf{FAD:} Table \ref{tab:few-shot ad} also shows the comparison of SOFS with other state-of-the-art few-shot anomaly detection approaches. Although SOFS only receives the image information, it can achieve a competitive performance with other multi-modal models. This shows that SOFS has strong capabilities for both FSS and FAD. 

In addition, we show the performance of FSS under domain shifts. We only test the model trained on VISION V1 \cite{bai2023vision} for Defect Spectrum \cite{yang2023defect} and AeBAD \cite{zhang2023industrial}, we verify the performance under different domain shifts. To avoid conflicts between test classes and training classes, we exclude VISION V1 from our test. Test defect types are 81. The results are shown in Table \ref{tab:extra val}. Although these datasets contain larger defects than VISION V1 and different domain shifts, SOFS still achieves the best performance. More quantitative and qualitative results are shown in Appendix.

\begin{table}[]
\scriptsize
\caption{Few-shot anomaly detection results on MVTec AD and VisA datasets. Results are listed as the average of 10 runs. The results for AnomalyGPT and WinCLIP are reported from the original paper. PatchCore is reproduced by us. $^\clubsuit$ denotes  the multi-modal model.}
	\label{tab:few-shot ad}
	\centering
	\begin{tabular}{@{}cccccccc@{}}
		\toprule
		\multirow{2}{*}{Setup} &
		\multirow{2}{*}{Method} &
		\multicolumn{3}{c}{MVTec-AD} &
		\multicolumn{3}{c}{VisA} \\ \cmidrule(l){3-8} 
		&&Image-AUC &Pixel-AUC &Pixel-F$_{1}$-max &Image-AUC &Pixel-AUC &Pixel-F$_{1}$-max \\ \midrule
		\multirow{4}{*}{1-shot} 
		& PatchCore \cite{roth2022towards} {\fontsize{5pt}{2pt}\selectfont CVPR' 2022} &83.0 $\pm$ 1.0 &92.9 $\pm$ 0.2 &45.7 $\pm$ 0.5 &78.2 $\pm$ 0.9 &96.0 $\pm$ 0.2 &28.4 $\pm$ 0.3 \\
		& WinCLIP$^\clubsuit$ \cite{jeong2023winclip} {\fontsize{5pt}{2pt}\selectfont CVPR' 2023} &93.1 $\pm$ 2.0 &95.2 $\pm$ 0.5 &\bd 55.9 $\pm$ 2.7 &83.8 $\pm$ 4.0 &96.4 $\pm$ 0.4 &\bd41.3 $\pm$ 2.3 \\ 
		& AnomalyGPT$^\clubsuit$ \cite{gu2024anomalygpt} {\fontsize{5pt}{2pt}\selectfont AAAI' 2024} &\bd 94.1 $\pm$ 1.1 &95.3 $\pm$ 0.1 &- &87.4 $\pm$ 0.8 &96.2 $\pm$ 0.1 &- \\ 
		& \cellcolor{grey1}SOFS (Ours) &\cellcolor{grey1} 93.3 $\pm$ 0.4 &\cellcolor{grey1} \bd96.3 $\pm$ 0.1 &\cellcolor{grey1} 54.8 $\pm$ 0.3 &\cellcolor{grey1} \bd88.0 $\pm$ 0.4 &\cellcolor{grey1} \bd96.6 $\pm$ 0.1 &\cellcolor{grey1} 35.9 $\pm$ 0.3 \\ \midrule
		\multirow{4}{*}{2-shot}
		& PatchCore \cite{roth2022towards} {\fontsize{5pt}{2pt}\selectfont CVPR' 2022} &86.2 $\pm$ 0.5 &93.9 $\pm$ 0.2 &47.4 $\pm$ 0.4 &81.9 $\pm$ 0.5 &96.7 $\pm$ 0.1 &30.2 $\pm$ 0.3 \\
		& WinCLIP$^\clubsuit$ \cite{jeong2023winclip} {\fontsize{5pt}{2pt}\selectfont CVPR' 2023} &94.4 $\pm$ 1.3 &96.0 $\pm$ 0.3 &\bd58.4 $\pm$ 1.7 &84.6 $\pm$ 2.4 &96.8 $\pm$ 0.3 &\bd43.5 $\pm$ 3.3 \\ 
		& AnomalyGPT$^\clubsuit$ \cite{gu2024anomalygpt} {\fontsize{5pt}{2pt}\selectfont AAAI' 2024} &\bd 95.5 $\pm$ 0.8 &95.6 $\pm$ 0.2&- &88.6 $\pm$ 0.7 &96.4 $\pm$ 0.1 &- \\
        &\cellcolor{grey1} SOFS (Ours) &\cellcolor{grey1} 94.5 $\pm$ 0.3 &\cellcolor{grey1} \bd96.7 $\pm$ 0.1 &\cellcolor{grey1} 56.6 $\pm$ 0.2 &\cellcolor{grey1} \bd89.9 $\pm$ 0.5 &\cellcolor{grey1} \bd96.9 $\pm$ 0.1 &\cellcolor{grey1} 37.7 $\pm$ 0.4 \\ \midrule
        \multirow{4}{*}{4-shot}
		&PatchCore \cite{roth2022towards} {\fontsize{5pt}{2pt}\selectfont CVPR' 2022} &89.3 $\pm$ 0.7 &94.9 $\pm$ 0.1 &48.8 $\pm$ 0.3 &83.7 $\pm$ 0.4 &97.2 $\pm$ 0.1 &31.6 $\pm$ 0.3 \\
		&WinCLIP$^\clubsuit$ \cite{jeong2023winclip} {\fontsize{5pt}{2pt}\selectfont CVPR' 2023} &95.2 $\pm$ 1.3 &96.2 $\pm$ 0.3 &\bd59.5 $\pm$ 1.8 &87.3 $\pm$ 1.8 &\bd97.2 $\pm$ 0.2 &\bd 47.0 $\pm$ 3.0 \\
		&AnomalyGPT$^\clubsuit$ \cite{gu2024anomalygpt} {\fontsize{5pt}{2pt}\selectfont AAAI' 2024} &\bd96.3 $\pm$ 0.3 &96.2 $\pm$ 0.1 &-&90.6 $\pm$ 0.7 &96.7 $\pm$ 0.1 &- \\
		&\cellcolor{grey1} SOFS (Ours) &\cellcolor{grey1} 95.8 $\pm$ 0.4 &\cellcolor{grey1} \bd97.1 $\pm$ 0.1 &\cellcolor{grey1} 58.3 $\pm$ 0.2  &\cellcolor{grey1} \bd92.0 $\pm$ 0.3 &\cellcolor{grey1} 97.0 $\pm$ 0.1 &\cellcolor{grey1} 39.4 $\pm$ 0.3 \\ \bottomrule
	\end{tabular}
\end{table}

\begin{table}[!t]
  \caption{FSS performance under domain shifts. We only test models trained on VISION V1 for Defect Spectrum \cite{yang2023defect} and AeBAD \cite{zhang2023industrial} and report 1-shot/5-shot results using the mIoU (\%). The results except for SegGPT (under ViT Large) are under DINO v2 Base. “$^{\star}$$^{\star}$”: NR+MNDL. “$^{\star}$”: NR. The results are the average of 10 runs.}
  \label{tab:extra val}
  \centering
  \scriptsize
  \begin{tabular}{ccccccccccc}
    \toprule
    & PFENet~\cite{tian2020prior} & PFENet$^{\star}$$^{\star}$ & HDMNet~\cite{peng2023hierarchical} & HDMNet$^{\star}$$^{\star}$ & SegGPT~\cite{wang2023seggpt} & SegGPT$^{\star}$ & SOFS~(Ours) \\ \hline
    AeBAD & 13.4/19.1 & 16.4/19.7 & 15.9/24.4 & 22.1/\bd29.8 & 9.3/12.3 &9.5/11.3 & \bd22.4/29.4 \\
    Defect Spectrum &23.8/25.6 &35.9/36.8 &31.8/39.4 &40.2/43.8 &38.3/44.2 &37.1/41.7 & \bd43.7/\bd47.5\\ \bottomrule
  \end{tabular}
\end{table}

\subsection{Ablation Study}

We report the ablation studies in this section to investigate the effectiveness of each component. All ablation experiments are conducted under VISION V1 1-shot setting with DINO v2 Base backbone.

\textbf{Small object enhancement designs ablation.} Table \ref{tab:ablation_components1} shows results regarding the effectiveness of small object enhancement designs. The baseline (line 1) uses the bilinear interpolation for $\hat{\textbf{\textit{M}}}^{\rm{s}}$ and the standard resizing procedure, its performance is extremely low. Then we add the prototype feature intensity map (PFIM), resulting in improved results. Since many defects account for small area proportions in F0 and F1, the performances of F0 and F1 are still low. Due to “Cylinder” with large defects in F2, the improvement for F2 is significant. Subsequently, we only add the non-resizing procedure (NR). Compared with line 2, this brings a greater improvement. However, the improvement in F1 is still not enough. This is mainly because there are many defects with small pixel areas in F1. NR cannot deal with this kind of problem. Finally, we use PFIM and NR to simultaneously improve the performance for defects with small pixel areas and small area proportions, resulting in a significant improvement for F1.

\textbf{Other components ablation.} Table \ref{tab:ablation_components2} shows results regarding the effectiveness of other components in SOFS. The baseline (line 1) is built upon the matching mechanism in \cite{peng2023hierarchical} (a type of cross attention), the learnable prediction classifier (a multi-layer perceptron) used in \cite{peng2023hierarchical, tian2020prior} and trained by Dice loss \cite{milletari2016v}. We first replace Dice loss with the proposed mixed normal Dice loss (MNDL), resulting in a large improvement for F0 and F1. However, since the training on F2 does not include large defects (“Cylinder” with large defects is in F2), the model with too many parameters causes overfitting to small defects. Then we replace the matching mechanism in \cite{peng2023hierarchical} with a non-learnable feature fusion, the performances for F0, F1, and F2 improve. Subsequently, we add an abnormal prior map (APM). As shown in Fig. \ref{fig_5}, APM can reduce the false positive regions of missing defects, resulting in further improvements. Finally, we replace the learnable prediction classifier with a meta prediction, bringing $1.5\%$ and $1.1\%$ mIoU improvement for F0 and F1. But this also brings a slight degradation for F2. We find that this is due to insufficient prediction for large defects.

\begin{table}[!t]
\begin{minipage}[t]{0.35\textwidth}
    \caption{Ablation studies (mIoU (\%)). “NR”: non-resizing procedure. “PFIM”: prototype feature intensity map.}
    \label{tab:ablation_components1}
    \centering
    \small
    \setlength\tabcolsep{3pt}
    \begin{tabular}{ccccc}
        \toprule
        NR & PFIM  & F0 & F1 & F2  \\
        \midrule
                       &            &    3.9  & 3.3   & 11.2     \\
             &    \checkmark   &   7.6 & 7.3  & 21.1         \\
        \checkmark     &            &   20.2 & 14.0        & \bd22.5       \\
        \checkmark     & \checkmark &   \bd22.9 & \bd23.1   & 21.6   \\
        \bottomrule
    \end{tabular}
\end{minipage}
\hfill
\begin{minipage}[t]{0.6\textwidth}
    \caption{Ablation studies (mIoU (\%)). “MNDL”: mixed normal Dice loss. “NLFF”: non-learnable feature fusion. “APM”: abnormal prior map. “MC”: meta prediction.}
    \label{tab:ablation_components2}
    \centering
    \small
    \setlength\tabcolsep{3pt}
    \begin{tabular}{ccccccc}
        \toprule
        MNDL & NLFF & APM & MC & F0 & F1 & F2  \\
        \midrule
                       &            &                  &        & 8.1 & 10.1         & 5.1         \\
        \checkmark     &            &                  &          & 18.1 & 18.9   & 11.1         \\
        \checkmark     & \checkmark &                  &         & 19.7 & 19.1   & 15.1        \\
        \checkmark     & \checkmark & \checkmark       &        & 21.4  & 22.0   & \bd22.2       \\
        \checkmark     & \checkmark & \checkmark       & \checkmark     & \bd22.9 & \bd23.1    & 21.6        \\
        \bottomrule
\end{tabular}
\end{minipage}
\end{table}

\begin{figure}[t]
	\centering
	\includegraphics[width=0.8\linewidth]{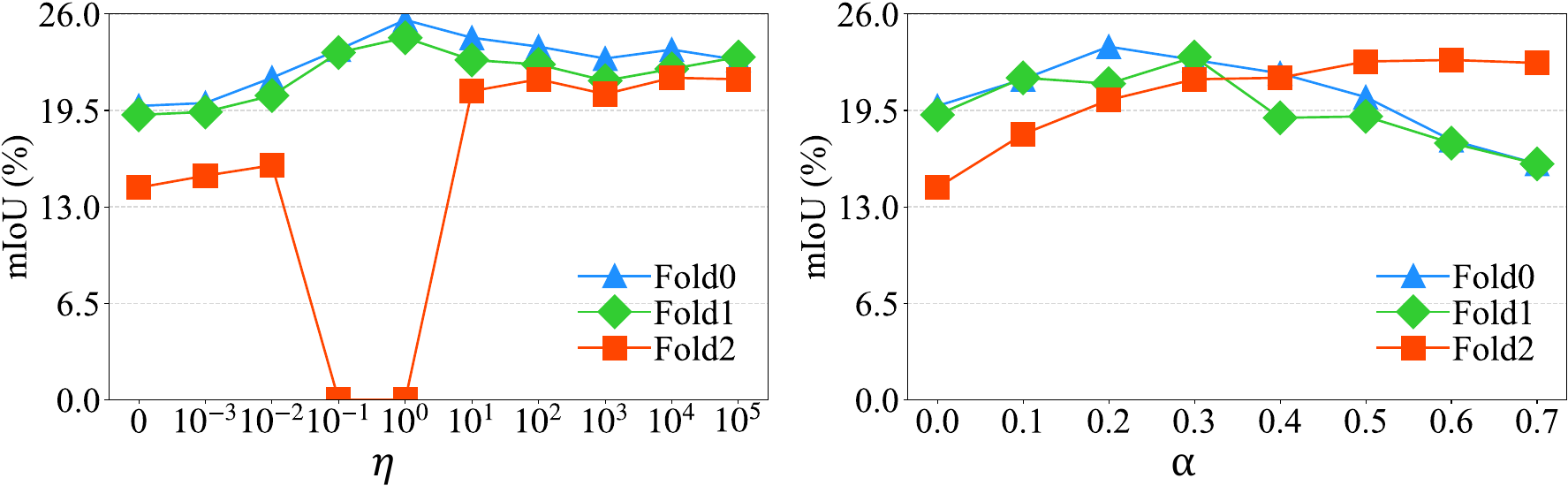}
	\caption{The results of different $\eta$ and $\alpha$ in the mixed normal Dice loss.}
	\label{fig_mndl}
\end{figure}

\textbf{$\eta$ and $\alpha$ in the mixed normal Dice loss.} Fig. \ref{fig_mndl} shows the results for different $\eta$ and $\alpha$. $\eta$: If $\eta$ is small ($0-10^{-2}$), it does not penalize false positives enough, resulting in a relatively weak performance, especially for Fold 2. If $\eta$ is set to $10^{-1}$ or $10^{0}$, the performance of Fold 0 and Fold 1 improve significantly. However, the performance of Fold 2 is 0. We found that after the training started, the model quickly fell into a local optimum with an output of all 0. This is mainly because the training for Fold 2 does not contain the large defects (Cylinder). In other words, if the training samples are all small defects, then some $\eta$ will cause the training to fall into a useless local optimal solution. If $\eta$ is set to a large number, this phenomenon will disappear. $\alpha$: Different $\alpha$ have a significant impact on the results. If $\alpha$ is small, the number of normal query samples per iteration is small (possibly none), resulting in no penalty for false positives. If $\alpha$ is large, it causes the underfitting for defective samples.

\begin{wrapfigure}{r}{0.5\linewidth}
    \centering
    \vspace{-15pt}
    \includegraphics[width=0.8\linewidth]{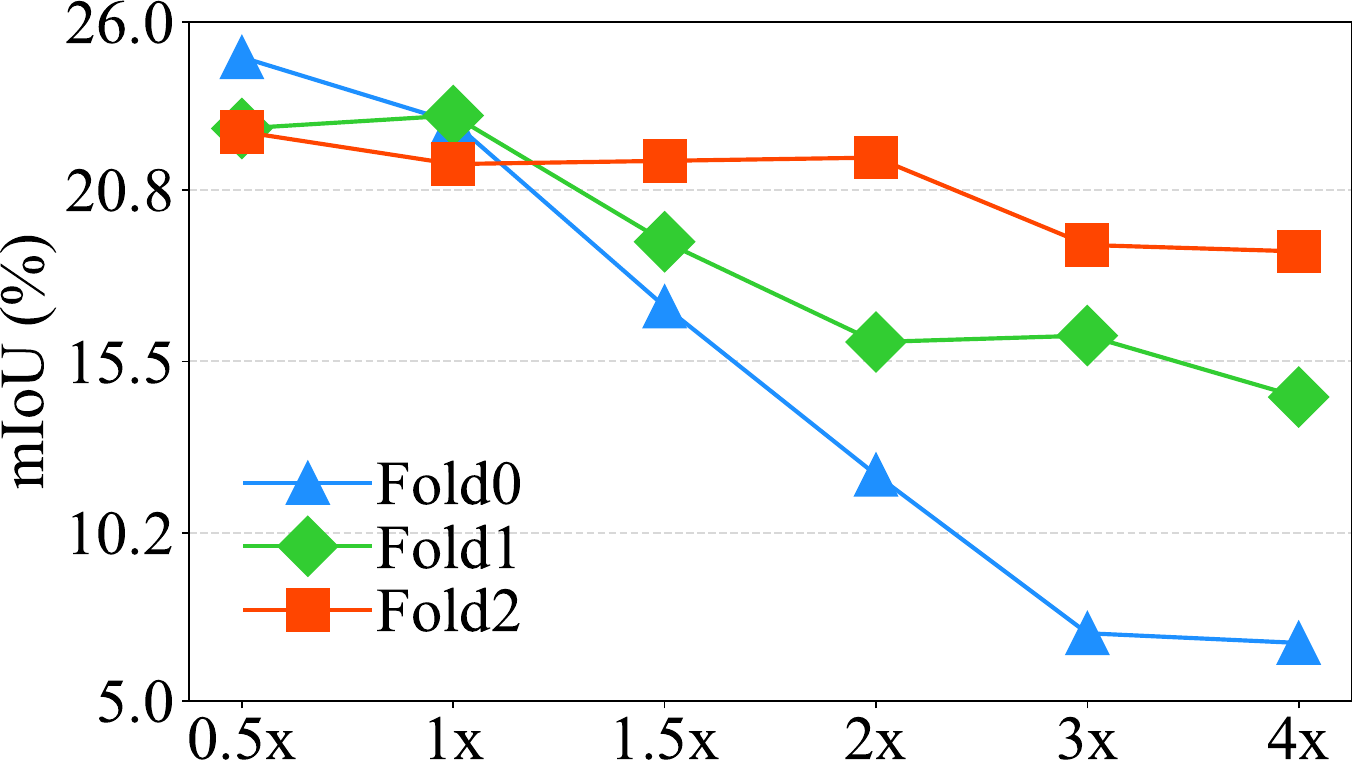}
    \vspace{-10pt}
    \caption{The results as the test input size evolves in the sliding window mechanism.}
    \label{fig_abl_1}
    \vspace{-15pt}
\end{wrapfigure}

\textbf{Input size in test.} Fig. \ref{fig_abl_1} shows the results as the test input size evolves in the sliding window mechanism (the stride is the same as the input size). We can find that different test input sizes have a significant impact on the segmentation of small objects, especially for the defects with small area proportions (Fold 0). The smaller the input size, the higher the latency of the model. We should choose the test input size based on actual requirements.

\section{Conclusion}
\label{5_conclusion}

In this paper, we propose SOFS to tackle the problems existing in supervised learning under a close-set setting and industrial anomaly detection in VII. SOFS significantly improves the small object segmentation by avoiding resizing and a prototype feature intensity downsampling. Further, we add an abnormal prior map in SOFS to guide the model to reduce false positives for backgrounds, while enabling SOFS to realize both FSS and FAD. In addition, we propose a mixed normal Dice loss, non-learnable feature fusion, and meta prediction to alleviate the overfitting problem. The experiments for different defects under an open-set setting demonstrate the superior performance of SOFS. We hope that SOFS can provide some insights for the community of VII.

\textbf{Limitations and Future Works} 1) As shown in Fig. \ref{fig_6}, the differences among defects in the same object are slight, which leads to some false positives. We think that test-time training/adaptation \cite{sun2020test, liang2023comprehensive} may be a promising method to tackle this problem. 2) As shown in Fig. \ref{vis: app_fig_2} (7th row) and Fig. \ref{vis: app_fig_5} (3rd row), SOFS produces false negatives for defects with extreme aspect ratio. This may originate from the receptive field of the encoder. 3) If SOFS is used to process high-resolution images, due to the sliding window mechanism in the test, the latency of SOFS is high. We will explore some designs to reduce latency and segment defects with extreme aspect ratios in the future. 4) SOFS is allowed to adopt a mixture of normal support images and defective support images for training and inference, which is a more common scenario in reality. We will explore this direction in the future.



{\small
\bibliographystyle{abbrvnat}
\bibliography{bibliography}
}


\appendix

\section{Appendix}

\subsection{Supplemental Details of SOFS}

\textbf{Masked average pooling.} The masked average pooling is formulated as follows:

\begin{equation}
\label{EN:8}
MaskAvgPool(\textbf{\textit{F}},\textbf{\textit{M}})= \frac{\rm{sum}^{\prime}(\textbf{\textit{F}} \odot \vartheta(\textbf{\textit{M}}))}{\rm{sum}(\textbf{\textit{M}})} ,
\end{equation}

where $\textbf{\textit{F}}\in \mathbb{R}^{H\times W\times C}$, $\textbf{\textit{M}} \in \mathbb{R}^{H\times W}$, $\vartheta(\cdot): \mathbb{R}^{H\times W} \mapsto \mathbb{R}^{H\times W\times C}$ refers to first expanding the new dimension and then replicating along the expanded dimension, $\rm{sum}^{\prime}(\cdot): \mathbb{R}^{H\times W\times C}\mapsto \mathbb{R}^{C}$ refers to the addition of spatial features, $\rm{sum}(\cdot): \mathbb{R}^{H\times W}\mapsto \mathbb{R}$ refers to the addition along spatial dimensions.  

\textbf{Extension to \textit{K}-shot setting.} In extension to \textit{K}-shot (\textit{K} > 1) setting, \textit{K} support images with their annoted masks $\textbf{\textit{S}}=\{{(\textbf{\textit{I}}^{\rm{s}}_{i}, \textbf{\textit{M}}^{\rm{s}}_{i})}\}^{K}_{i=1}$ and the query set $\textbf{\textit{Q}}=\{ {(\textbf{\textit{I}}^{\rm{q}}, \textbf{\textit{M}}^{\rm{q}})}\}$ are given. $\hat{\textbf{\textit{M}}}^{\rm{s}}_{i}$, $\textbf{p}_{i}$, $\textbf{\textit{M}}^{\rm{q}_{i}}_{\rm{s}}$ and $\textbf{\textit{M}}^{\rm{q}_{i}}_{\rm{a}}$ are firstly calculated, where $\textbf{p}_{i}$ is extracted by a $MaskAvgPool(\textbf{\textit{F}}_{i}^{\rm{s}},\hat{\textbf{\textit{M}}}^{\rm{s}})$, $\textbf{\textit{M}}^{\rm{q}_{i}}_{\rm{s}}$ denote a semantic prior map calculated by $\textbf{\textit{F}}_{i}^{\rm{s}}$ and $\textbf{\textit{F}}^{\rm{q}}$. Then we stack all $\textbf{\textit{M}}^{\rm{q}_{i}}_{\rm{s}}$ and $\textbf{\textit{M}}^{\rm{q}_{i}}_{\rm{a}}$ along the number dimension of support set respectively. And we calculate the mean of maps concatenated toghter to obtain $\textbf{\textit{M}}^{\rm{q}}_{\rm{s}}$ and $\textbf{\textit{M}}^{\rm{q}}_{\rm{a}}$. In the feature fusion, we calculate the cosine similarity between the query feature and all support features. We concat all $\textbf{\textit{F}}_{i}^{\rm{s}}$ along the spatial dimension. The remaining steps are the same as 1-shot.

\textbf{\textit{K}-shot implementation.} Due to the non-resizing procedure, we can sample multiple support images under a 1-shot scenario. These support images are from the random crops containing defects. In 1-shot experiments, we sample 4 random crops in one support image. In 5-shot experiments, we sample 1 random crop in one support image. We use this implementation for all methods.

\textbf{Few-shot Anomaly Detection.} We use the maximum value on $\hat{\textbf{\textit{M}}}^{\rm{q}}_{\rm{pred}}$ to indicate the abnormal degree of the image.

\subsection{Supplemental Implementation Details}

\textbf{SOFS.} SOFS is built upon the Pytorch \cite{paszke2019pytorch} framework. All models are trained and tested on 8 NVIDIA GeForce RTX 2080 Ti GPUs. All reimplementation methods and SOFS use the same training augmentations, including rotation, blur, and flip. In the non-resizing procedure, we use the random crop, where the crop image includes the defect. We only use the non-resizing procedure if the area proportion of defect in the support set is less than $1\%$, otherwise, we use the common resizing. In addition, following the previous works, SOFS adopts features from different layers of backbones to further improve the useful information. 

SOFS is trained in an episode fashion for 50 epochs for VISION V1, the batch size is 4 for every GPU. During training, AdamW optimizer is adopted, and the learning rate is set to $1\rm{e}^{-5}$, the weight decay is 0.01, and the “poly” strategy is used to adjust the learning rate. During testing, predictions are resized back to the original sizes of the input images, keeping the groundtruth labels intact.

In the few-shot anomaly detection, SOFS adopts the common resizing procedure.

\textbf{Reimplementation of Comparison Methods} We compare SOFS with the recent methods: PFENet \cite{tian2020prior}, HDMNet \cite{peng2023hierarchical}, and SegGPT \cite{wang2023seggpt}. We reproduce them following the official codes and adjust the hyperparameters to maximize performance. For PFENet and HDMNet, we train the models by cross entropy loss and Dice loss \cite{milletari2016v}. Due to the small objects, we adjust the downsampling modules in PFENet and HDMNet, we do not downsample the feature map. The input sizes of models are the same as SOFS. In addition, we do not mask the support features in the feature extraction of the freezing encoder like previous methods, masking leads to poor performance. We only test the model for SegGPT, and the input size is set to $448\times 448$, which is the same as the original setting. We do not observe the improvement if the input size is set to $512\times 512$.

We use ResNet-50 \cite{he2016deep} and DINO v2 Base \cite{oquab2023dinov2} as the encoder to extract features with freezing parameters to verify the effectiveness of the proposed method and reimplementation methods on different backbones. For ResNet-50, we use the multi-scale features from layer 2 and layer 3, and the feature processing method in \cite{roth2022towards} is adopted to upsample the high-level features. We do not use PSPNet \cite{zhao2017pyramid} to train the base learner, which is used in the previous methods \cite{peng2023hierarchical, lang2023base}. We find that this causes a severe overfitting problem to fit the category-specific distribution. Using the original pretrained model can alleviate this. The feature dimension $C$ is set to 64 and $C_{1}$ is set to 64. For DINO v2 Base, we use the features from layers 5, 6, 7, 8, 9, 10. The feature dimension $C$ is set to 256 and $C_{1}$ is set to 256. To ensure fairness, the features from the different layers of backbones for SOFS and reimplementation methods are the same. Since the backbone cannot be replaced, SegGPT is verified by ViT-Large \cite{dosovitskiy2020image}.

\begin{figure}[t]
	\centering
	\includegraphics[width=\linewidth]{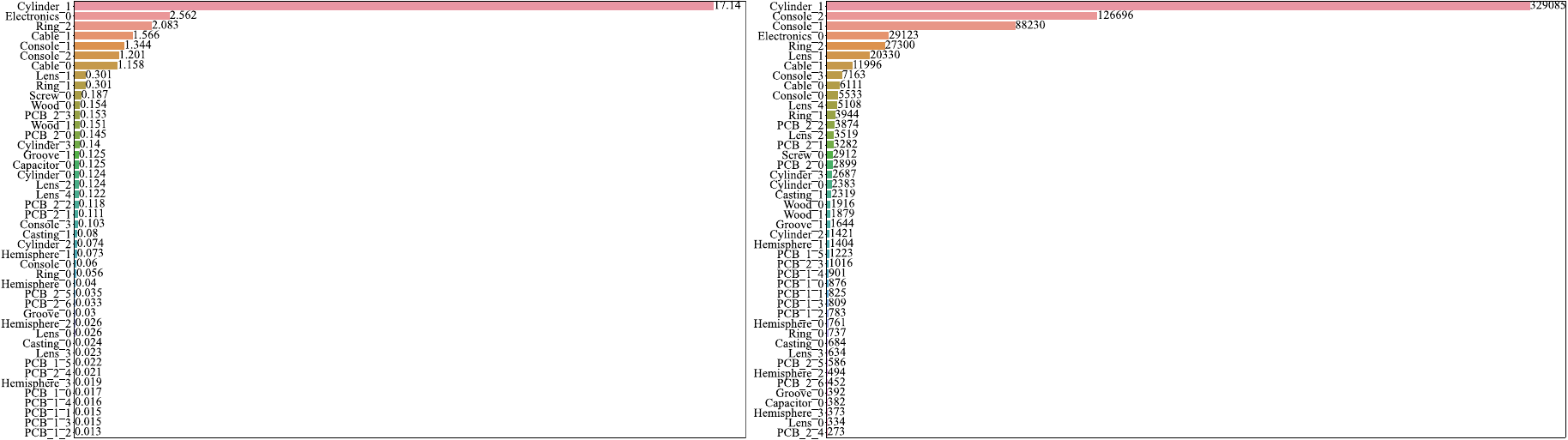}
	\caption{Area proportions (\%) (left) and pixel areas (number of pixels) (right) for different defects on VISION V1.}
	\label{supp_fig_1}
\end{figure}

\begin{table}
  \begin{center}
    \caption{Test objects in each fold on VISION V1.}
  \label{tab: supp_table_1}
    \begin{tabular}{cccccc}
    \toprule
    Fold 0 & PCB\_1 & Hemisphere & Wood & Ring & Cable \\
    Fold 1 & Casting & Capacitor & PCB\_2 & Electronics & Console \\
    Fold 2 & Groove & Lens & Screw & Cylinder & \\ \bottomrule
    \end{tabular}
  \end{center}
\end{table}

In the experiments on VISION V1 and Defect Spectrum, we train the models with 3 runs for PFENet, HDMNet, and SOFS, and then test each model with 5 runs, for a total of 15 runs. For SegGPT, we only test it with 15 runs. In the experiments of domain shifts (see Section \ref{supp_es}), we test models with 10 runs. To ensure performance stability and fairness for comparison, the query/support pairs sampling by all methods are the same (the non-resizing procedure adopts the crops of the image).

\subsection{Supplemental Details for Dataset}

\textbf{VISION V1.} VISION V1 \cite{bai2023vision} released in CVPR 2023 Vision-based InduStrial InspectiON workshop, is one of the largest and most diverse manufacturing datasets in terms of both image number and available annotations. VISION V1 consists of 14 objects sourced from Roboflow and \cite{ding2019tdd}, 10k high-quality defect segmentation annotations on 4k images, spanning 44 defect types. Area proportions and pixel areas (number of pixels) for different defects are shown in Fig. \ref{supp_fig_1}. $84\%$ of defect types have area proportions less than $0.3\%$, and $38\%$ of defect types have pixel areas less than $900$. These characteristics bring great challenges to defect segmentation. In the few-shot semantic segmentation, we divide all objects into 3 folds, we ensure that the distributions of area proportions of defects in each fold are roughly the same. The test objects for each fold are shown in Table \ref{tab: supp_table_1}.

\begin{figure}[t]
	\centering
	\includegraphics[width=\linewidth]{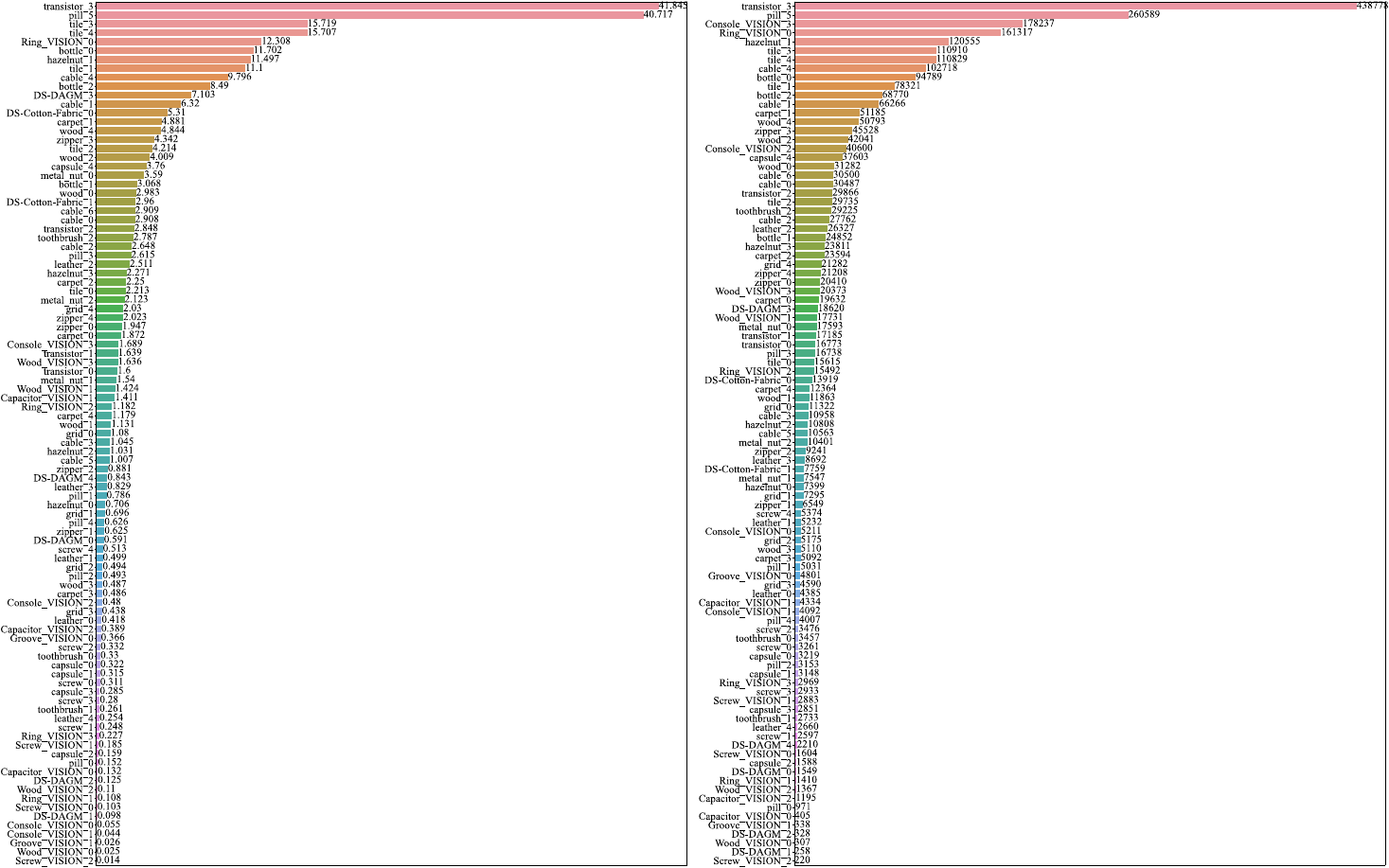}
	\caption{Area proportions (\%) (left) and pixel areas (number of pixels) (right) for different defects on Defect Spectrum.}
	\label{supp_fig_2}
\end{figure}

\textbf{Defect Spectrum.} Defect Spectrum \cite{yang2023defect} relabels the data from MVTec AD \cite{bergmann2019mvtec}, DAGM2007 \cite{wieler2007weakly}, part of VISION V1 \cite{bai2023vision} and Cotton-Fabric \cite{fabric}, ensuring that every image is accompanied by precise and diverse category annotations. It is the dataset with the largest number of defect categories so far, including in total 97 defect types. Area proportions and pixel areas (number of pixels) for different defects are shown in Fig. \ref{supp_fig_2}. $63\%$ of defect types have area proportions less than $2\%$. In the few-shot semantic segmentation, we divide all objects into 3 folds, we ensure that the distributions of area proportions of defects in each fold are roughly the same. The test objects for each fold are shown in Table \ref{tab: supp_table_2}. Compared with the defects in VISION V1, the defects in Defect Spectrum are larger. The objects in this dataset are more under experimental conditions, \ie, placing a single object in a noise-free background, and there is a certain gap between the defects that need to be detected in real manufacturing. Thus, we choose VISION V1 as the main experimental dataset. 

\begin{table}
  \begin{center}
    \caption{Test objects in each fold on Defect Spectrum.}
     \label{tab: supp_table_2}
    \tiny
    \begin{tabular}{cccccccccc}
    \toprule
    Fold 0 & Screw\_VISION & Ring\_VISION & Capacitor\_VISION & toothbrush & grid & cable & metal\_nut & tile & \\
    Fold 1 & Wood\_VISION & Console\_VISION & pill & leather & carpet & hazelnut & transistor & \\
    Fold 2 & Groove\_VISION & DS-DAGM & capsule & screw & wood & zipper & DS-Cotton-Fabric & bottle \\ \bottomrule
    \end{tabular}
  \end{center}
\end{table}

\textbf{AeBAD.} AeBAD \cite{zhang2023industrial} is used to detect defects in aero-engine blades. The defects of blades are under different domains, including different illuminations, backgrounds, and views. The defects contain 4 types, involving some minor defects. This dataset is used to verify the robustness of different models for different domains. In \ref{supp_es}, we only test the model trained on VISION V1 \cite{bai2023vision} for Defect Spectrum \cite{yang2023defect} and AeBAD \cite{zhang2023industrial}, we verify the performance under domain shifts.

\subsection{Supplemental Experiments}
\label{supp_es}

The results (mean IoU) of different defect types of every object on VISION V1 are shown in Table \ref{tab:vision detail 1 shot} and \ref{tab:vision detail 5 shot}. SOFS achieves the best in most cases.

\begin{table}[!t]
  \caption{The results (mean IoU (\%)) of different defect types of every object on VISION V1. The results except for SegGPT (under ViT Large) are under DINO v2 Base and 1-shot setting. “$^{\star}$$^{\star}$”: non-resizing procedure+mixed normal Dice loss. “$^{\star}$”: non-resizing procedure. The results are the average of 15 runs.}
  \label{tab:vision detail 1 shot}
  \centering
  \scriptsize
  \begin{tabular}{ccccccccccc}
    \toprule
    & PFENet~\cite{tian2020prior} & PFENet$^{\star}$$^{\star}$ & HDMNet~\cite{peng2023hierarchical} & HDMNet$^{\star}$$^{\star}$ & SegGPT~\cite{wang2023seggpt} & SegGPT$^{\star}$ & SOFS~(Ours) \\ \hline
    PCB\_1 & 1.5 &4.2 &2.1 &24.2 &0.1 &9.3 & \textbf{32.4}\\
    Hemisphere &7.1 &3.5 &6.1 &9.5 &0.1 &8.9 & \textbf{14.1}\\
    Wood &7.5 &17.8 &12.6 &\textbf{20.0} &7.6 &3.0 & 16.6\\
    Ring &3.8 &4.0 &10.0 &10.6 &6.8 &7.8 & \textbf{13.4}\\
    Cable &25.1 &28.7 &23.2 &29.8 &6.4 &7.3 & \textbf{32.5}\\
    Casting &6.0 &8.3 &9.2 &13.3 &11.1 &10.9 & \textbf{17.1}\\
    Capacitor &10.1 &8.3 &13.3 &11.3 &11.5 &7.4 & \textbf{16.0}\\
    PCB\_2 &0.0 &2.5 &1.1 &15.2 &8.0 &16.0 & \textbf{29.3}\\
    Electronics &5.0 &15.1 &11.1 &24.2&8.1 &9.8 & \textbf{26.8}\\
    Console &6.5 &11.6 &9.5 &15.3 &14.9 &6.5 & \textbf{16.3}\\
    Groove &3.7 &3.5 &3.4 &4.8 &3.1 &2.5 & \textbf{6.9} \\
    Lens &13.8 &9.6 &9.7 &10.3 &11.8 &7.7 & \textbf{13.5}\\
    Screw &38.7 &38.1 &33.4 &23.8 &18.9 &27.5 & \textbf{44.2}\\
    Cylinder &10.9 &6.1 &10.0 &14.7 &38.9 &\textbf{39.4} & 33.5\\
    \bottomrule
  \end{tabular}
\end{table}

\begin{table}[!t]
  \caption{The results (mean IoU (\%)) of different defect types of every object on VISION V1. The results except for SegGPT (under ViT Large) are under DINO v2 Base and 5-shot setting. “$^{\star}$$^{\star}$”: non-resizing procedure+mixed normal Dice loss. “$^{\star}$”: non-resizing procedure. The results are the average of 15 runs.}
  \label{tab:vision detail 5 shot}
  \centering
  \scriptsize
  \begin{tabular}{ccccccccccc}
    \toprule
    & PFENet~\cite{tian2020prior} & PFENet$^{\star}$$^{\star}$ & HDMNet~\cite{peng2023hierarchical} & HDMNet$^{\star}$$^{\star}$ & SegGPT~\cite{wang2023seggpt} & SegGPT$^{\star}$ & SOFS~(Ours) \\ \hline
    PCB\_1 & 1.6 &5.0 &3.6 &32.6 &0.0 &11.1 & \textbf{39.4}\\
    Hemisphere &7.7 &5.8 &6.8 &8.5 &0.1 &10.4 & \textbf{14.1}\\
    Wood &9.6 &17.6 &13.6 &\textbf{20.9} &8.2 &3.6 & 19.4\\
    Ring &4.4 &4.9 &17.3 &19.1 &6.2 &7.1 & \textbf{19.9}\\
    Cable &25.7 &30.1 &25.7 &30.2 &12.7 &10.6 & \textbf{35.7}\\
    Casting &13.5 &8.7 &11.0 &15.6 &12.9 &16.4 & \textbf{18.2}\\
    Capacitor &9.7 &10.3 &18.2 &17.1 &18.2 &13.9 & \textbf{20.0}\\
    PCB\_2 &0.0 &2.6 &8.4 &18.2 &9.8 &19.6 & \textbf{34.1}\\
    Electronics &5.1 &14.5 &14.6 &\textbf{27.2}&9.0 &7.2 & 26.0\\
    Console &6.8 &11.4 &12.8 &\textbf{20.2} &18.2 &16.4 & 18.5\\
    Groove &4.0 &3.4 &5.1 &6.3 &2.4 &4.4 & \textbf{8.4} \\
    Lens &14.9 &11.3 &11.4 &11.0 &\textbf{18.6} &11.6 & 14.3\\
    Screw &38.1 &\textbf{44.1} &34.3 &12.1 &20.0 &34.6 & 29.1\\
    Cylinder &11.3 &7.3 &22.3 &24.9 &41.9 &\textbf{46.7} & 33.1\\
    \bottomrule
  \end{tabular}
\end{table}

\begin{figure}[t]
	\centering
	\includegraphics[width=1.\linewidth]{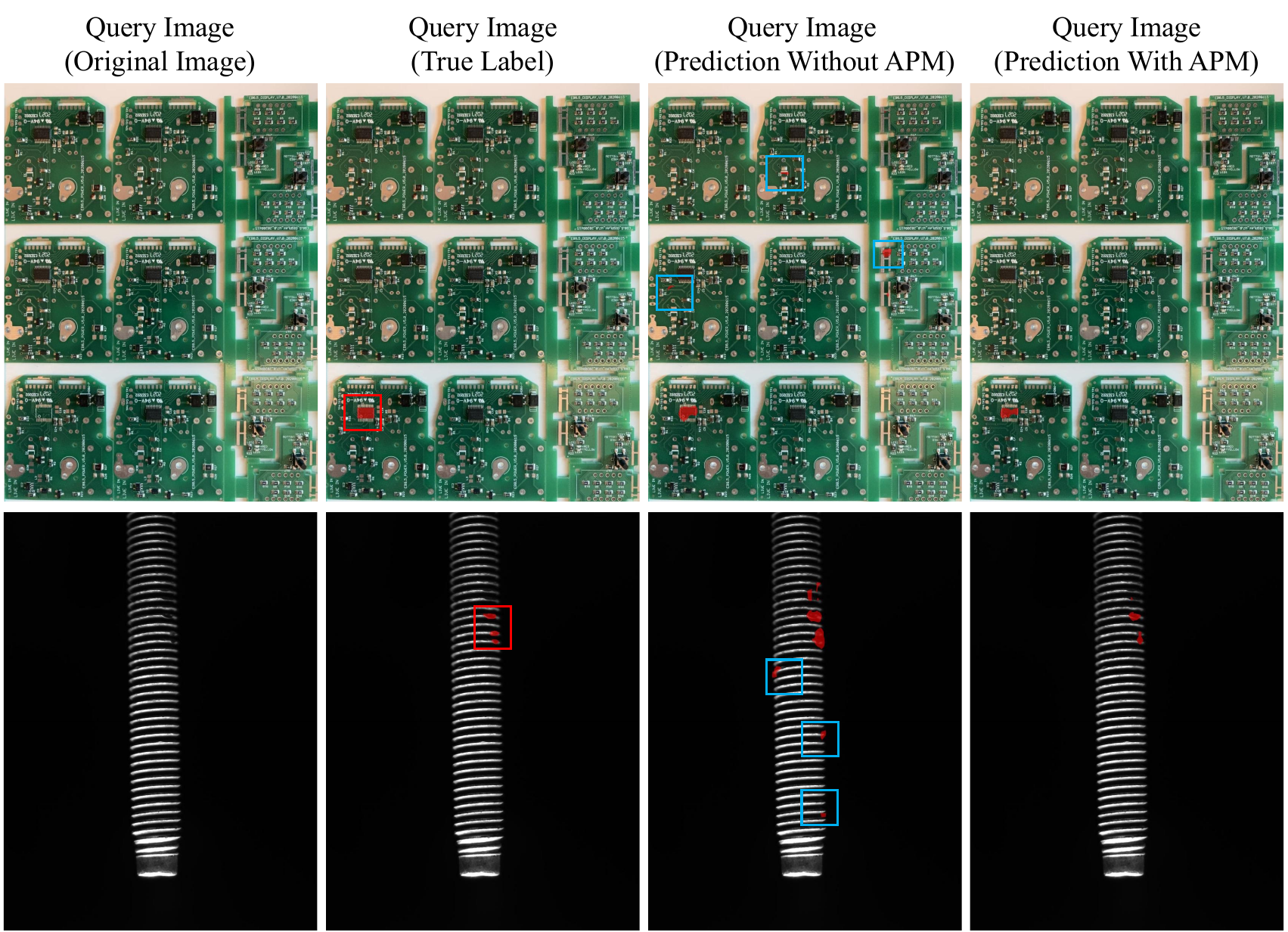}
	\caption{The outputs of the model with and without the abnormal prior map (APM). The support images are the same and the training random seeds are also the same. The blue solid line box indicates false positive regions.}
	\label{fig_5}
\end{figure}

\subsection{Supplemental Ablation Study}

The outputs of the model with and without the abnormal prior map are shown in Fig. \ref{fig_5}. We can find that APM can reduce the false positive regions of missing defects.

\subsection{More Qualitative Results}

\begin{figure}[t]
	\centering
	\includegraphics[width=0.85\linewidth]{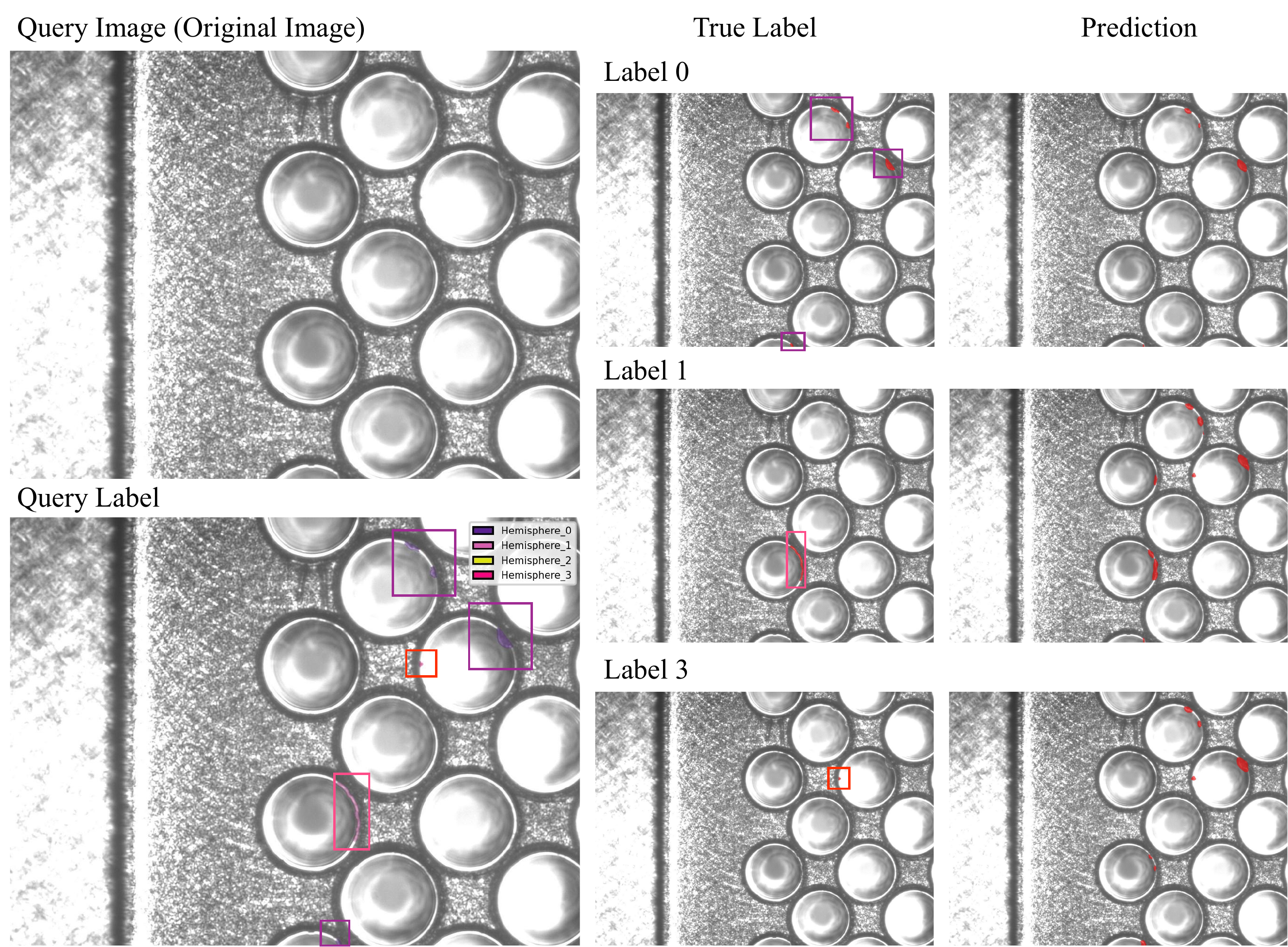}
	\caption{Different defects of a query image conditioned on different support images.}
	\label{fig_6}
\end{figure}

\begin{figure}[t]
	\centering
	\includegraphics[width=0.65\linewidth]{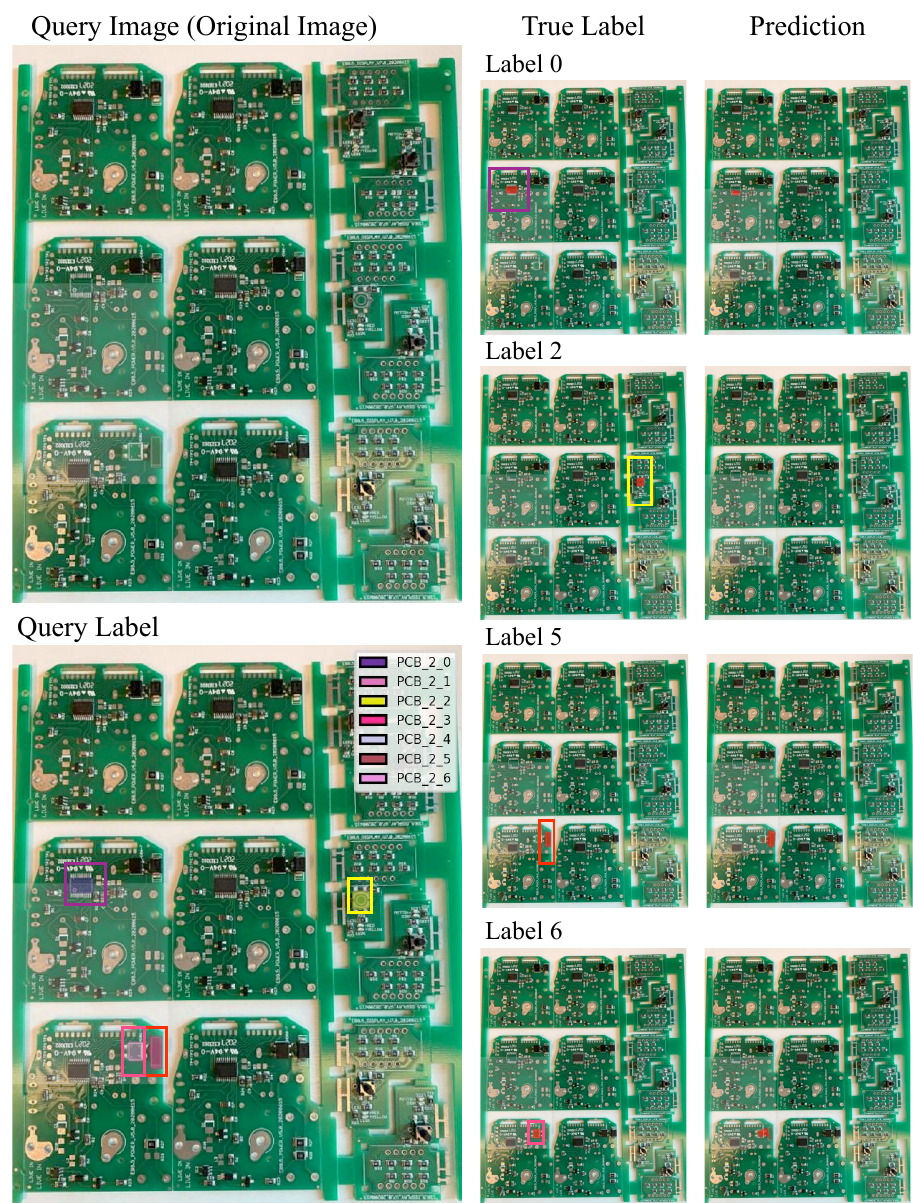}
	\caption{Different defects of a query image conditioned on different support images.}
	\label{fig_7}
\end{figure}

\begin{figure}[t]
	\centering
	\includegraphics[width=\linewidth]{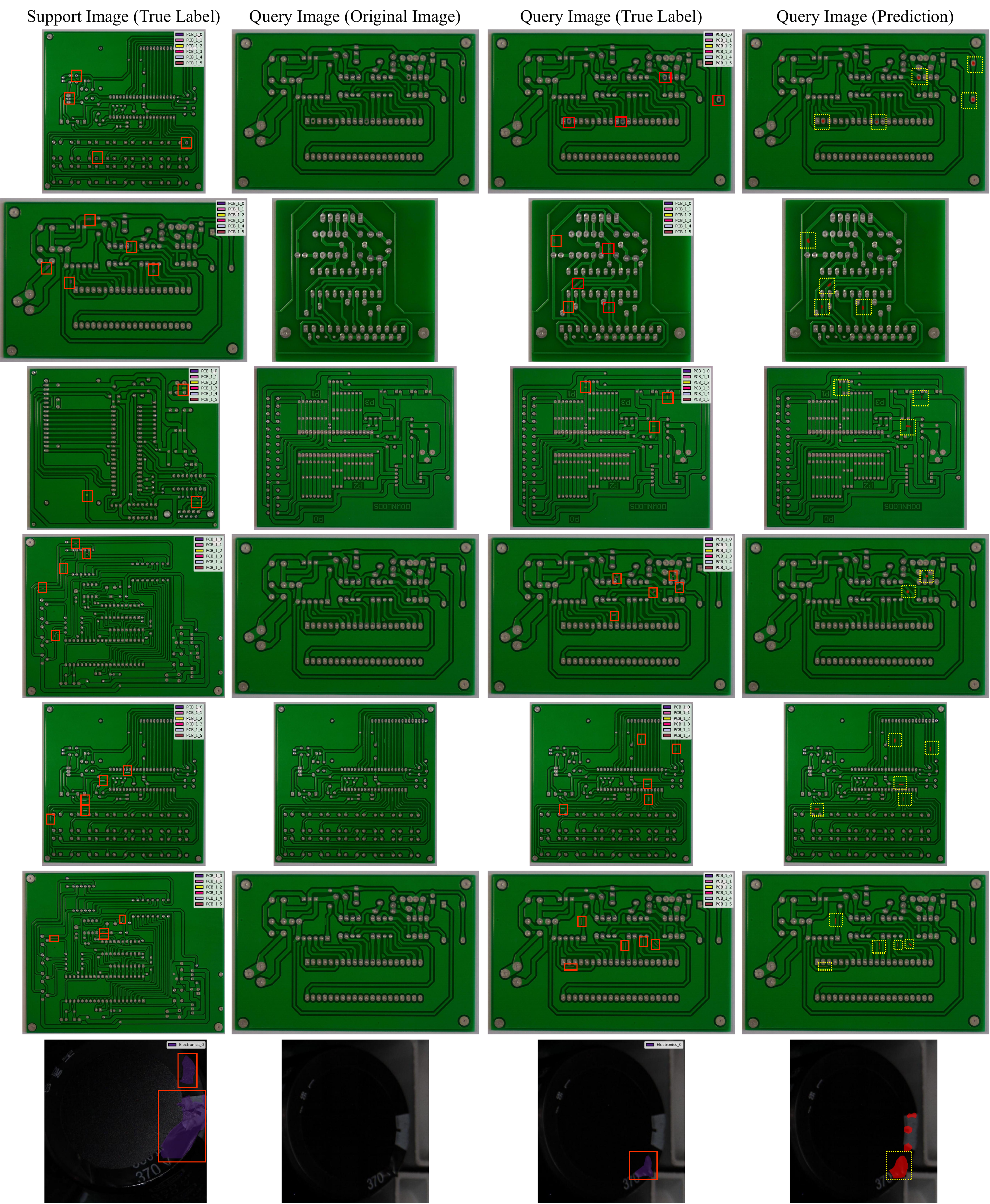}
	\caption{Qualitative results of SOFS under 1-shot setting. Corresponding to PCB\_1\_0 (the first class of defects of PCB\_1), PCB\_1\_1, PCB\_1\_2, PCB\_1\_3, PCB\_1\_4, PCB\_1\_5 and Electronics\_0 from top to bottom. Red solid line box indicates true labels, and yellow dashed line box indicates predictions. Due to the small defects, please use the electronic version to enlarge defects for easier viewing.}
	\label{vis: app_fig_1}
\end{figure}

\begin{figure}[t]
	\centering
	\includegraphics[width=\linewidth]{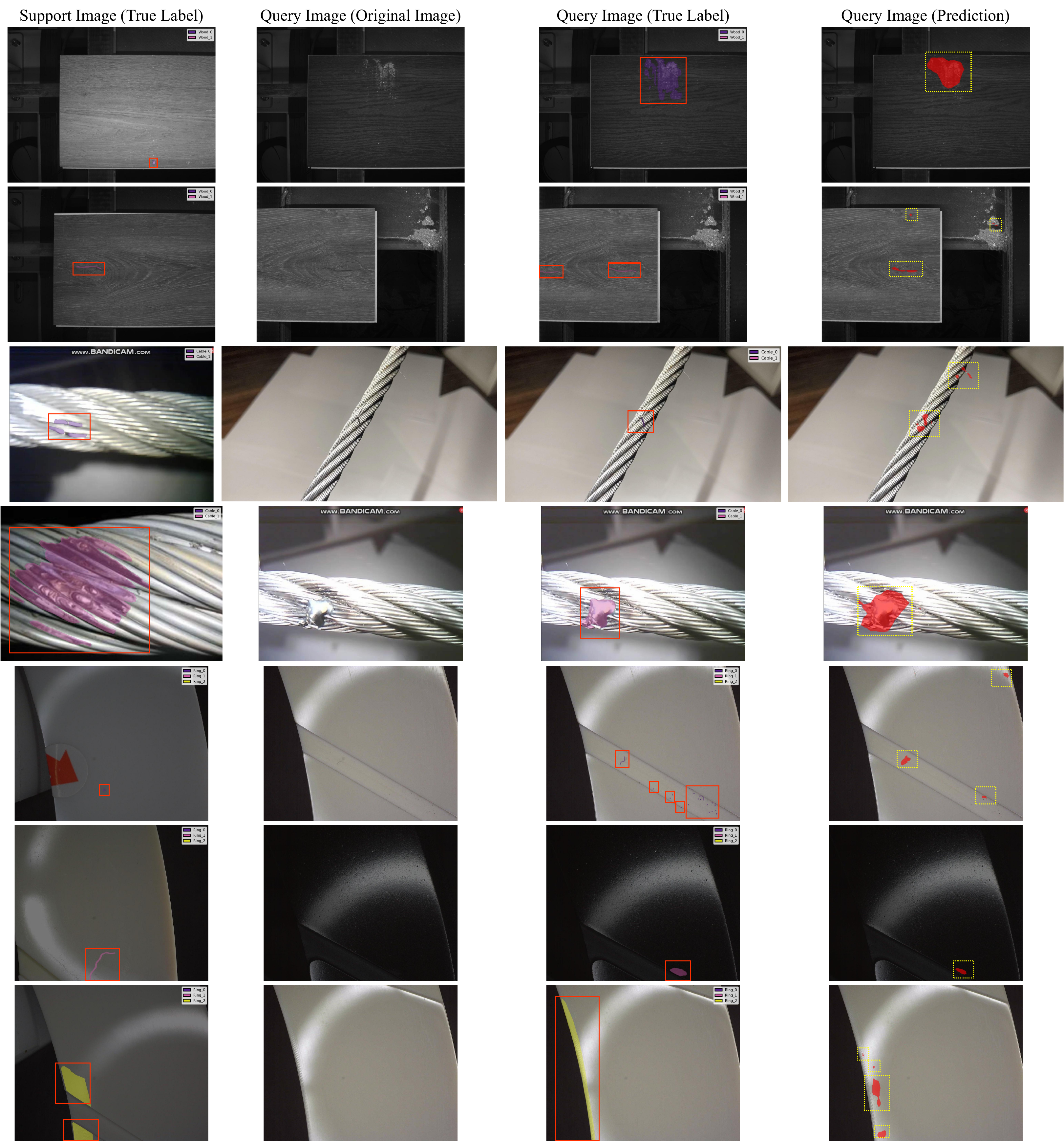}
	\caption{Qualitative results of SOFS under 1-shot setting. Corresponding to Wood\_0, Wood\_1, Cable\_0, Cable\_1, Ring\_0, Ring\_1 and Ring\_2 from top to bottom. Red solid line box indicates true labels, and yellow dashed line box indicates predictions. Due to the small defects, please use the electronic version to enlarge defects for easier viewing.}
	\label{vis: app_fig_2}
\end{figure}

\begin{figure}[t]
	\centering
	\includegraphics[width=\linewidth]{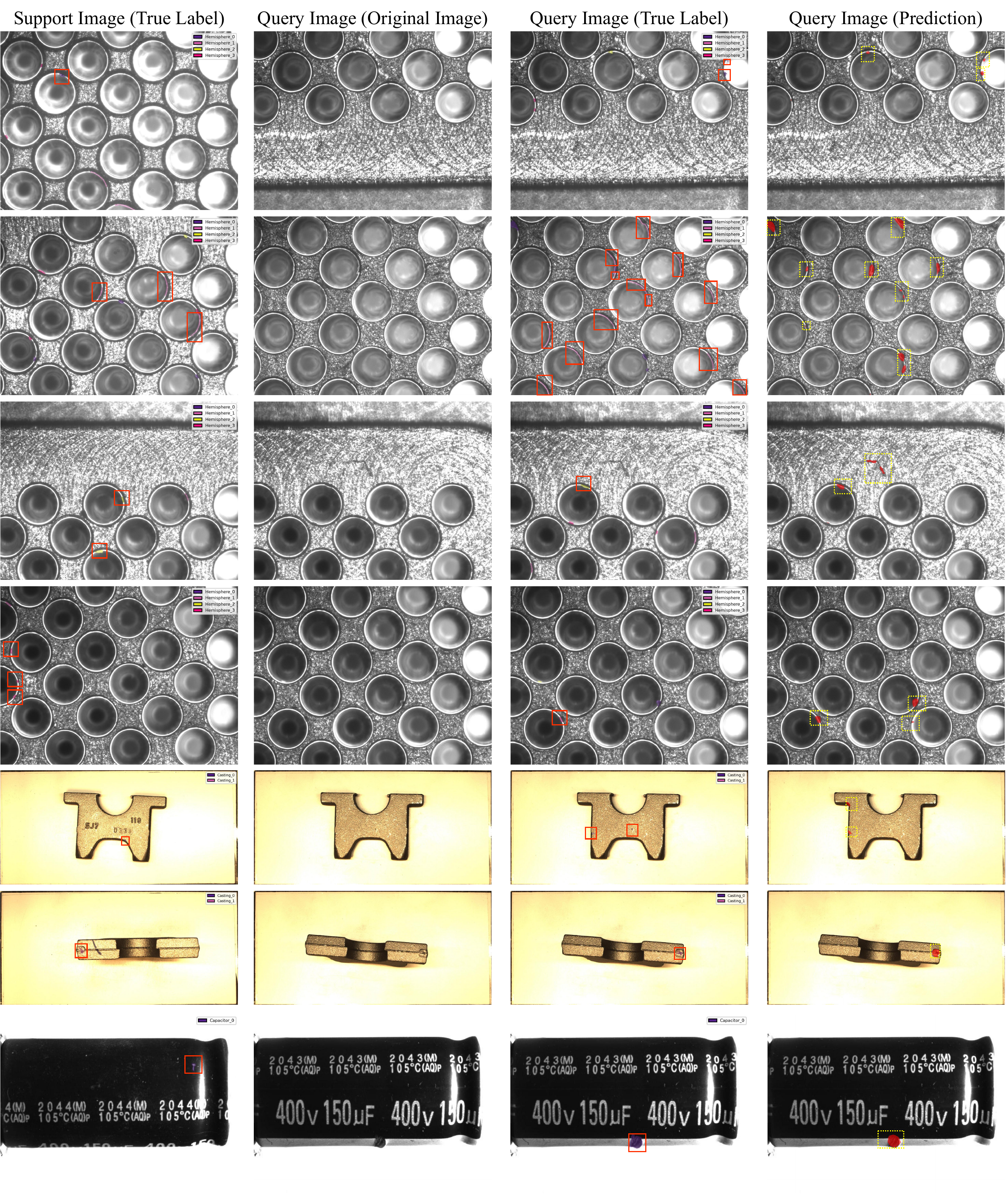}
	\caption{Qualitative results of SOFS under 1-shot setting. Corresponding to Hemisphere\_0, Hemisphere\_1, Hemisphere\_2, Hemisphere\_3, Casting\_0, Casting\_1 and Capacitor\_0 from top to bottom. Red solid line box indicates true labels, and yellow dashed line box indicates predictions. Due to the small defects, please use the electronic version to enlarge defects for easier viewing.}
	\label{vis: app_fig_3}
\end{figure}

\begin{figure}[t]
	\centering
	\includegraphics[width=0.7\linewidth]{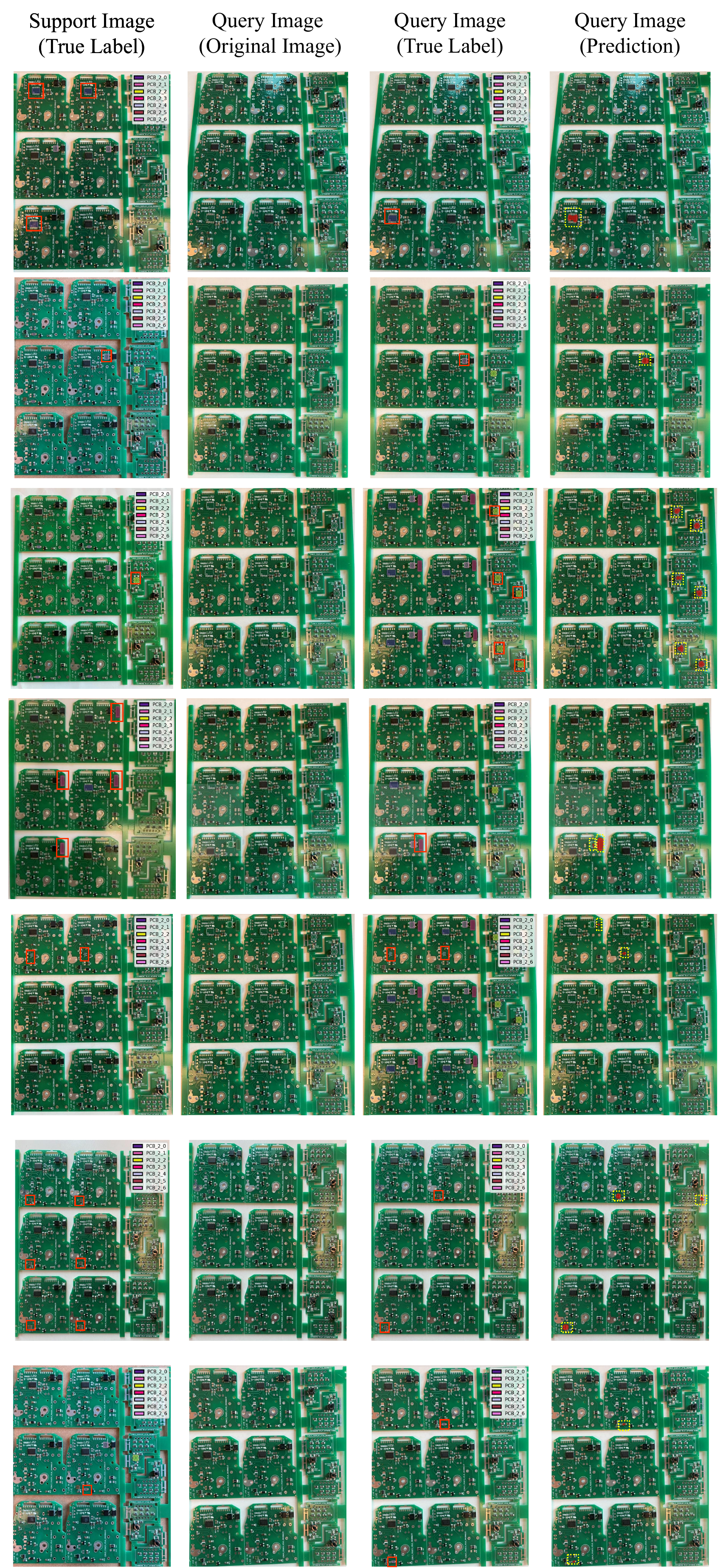}
	\caption{Qualitative results of SOFS under 1-shot setting. Corresponding to PCB\_2\_0, PCB\_2\_1, PCB\_2\_2, PCB\_2\_3, PCB\_2\_4, PCB\_2\_5 and PCB\_2\_6 from top to bottom. Red solid line box indicates true labels, and yellow dashed line box indicates predictions. Due to the small defects, please use the electronic version to enlarge defects for easier viewing.}
	\label{vis: app_fig_4}
\end{figure}

\begin{figure}[t]
	\centering
	\includegraphics[width=\linewidth]{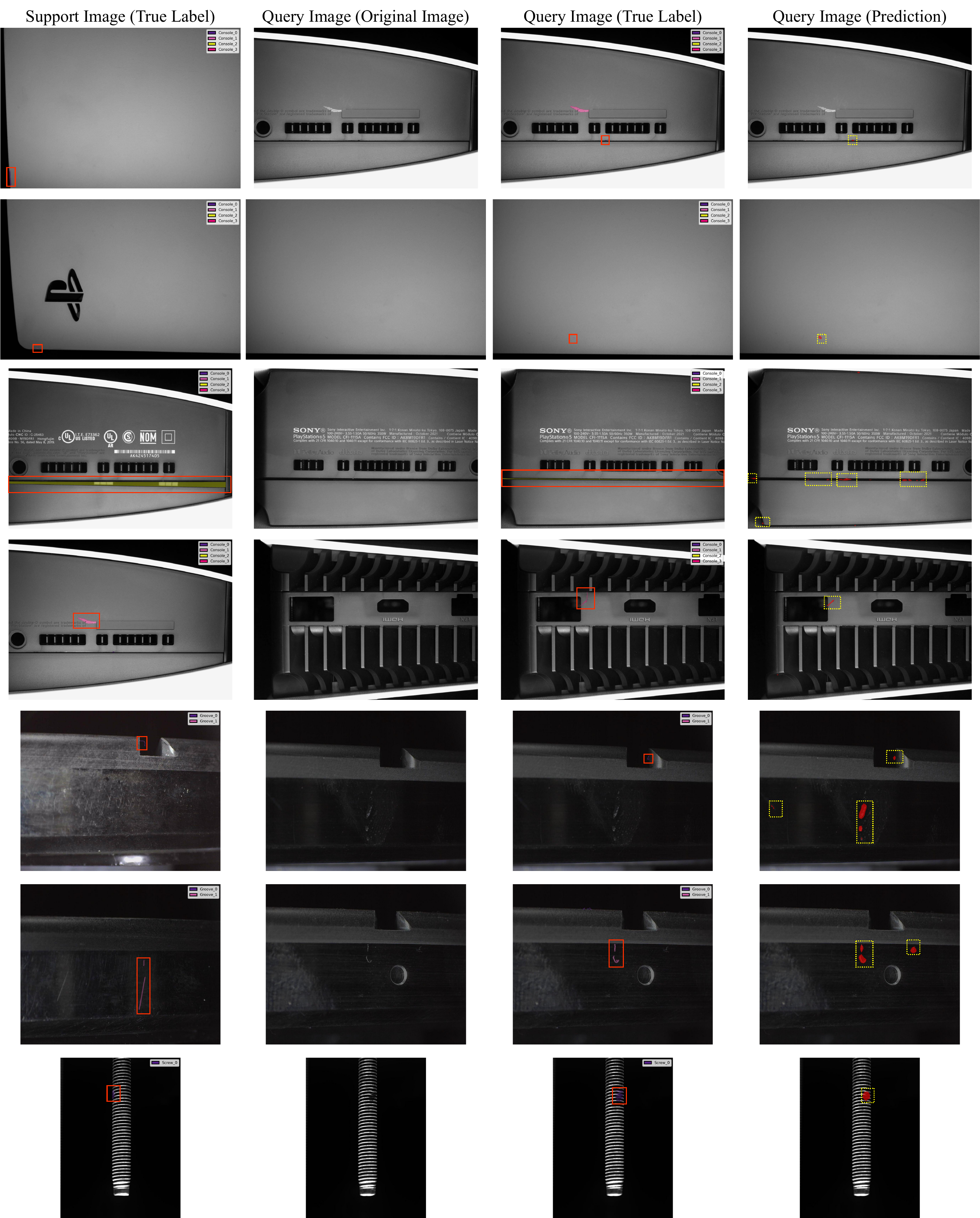}
	\caption{Qualitative results of SOFS under 1-shot setting. Corresponding to Console\_0, Console\_1, Console\_2, Console\_3, Groove\_0, Groove\_1 and Screw\_0 from top to bottom. Red solid line box indicates true labels, and yellow dashed line box indicates predictions. Due to the small defects, please use the electronic version to enlarge defects for easier viewing.}
	\label{vis: app_fig_5}
\end{figure}

\begin{figure}[t]
	\centering
	\includegraphics[width=\linewidth]{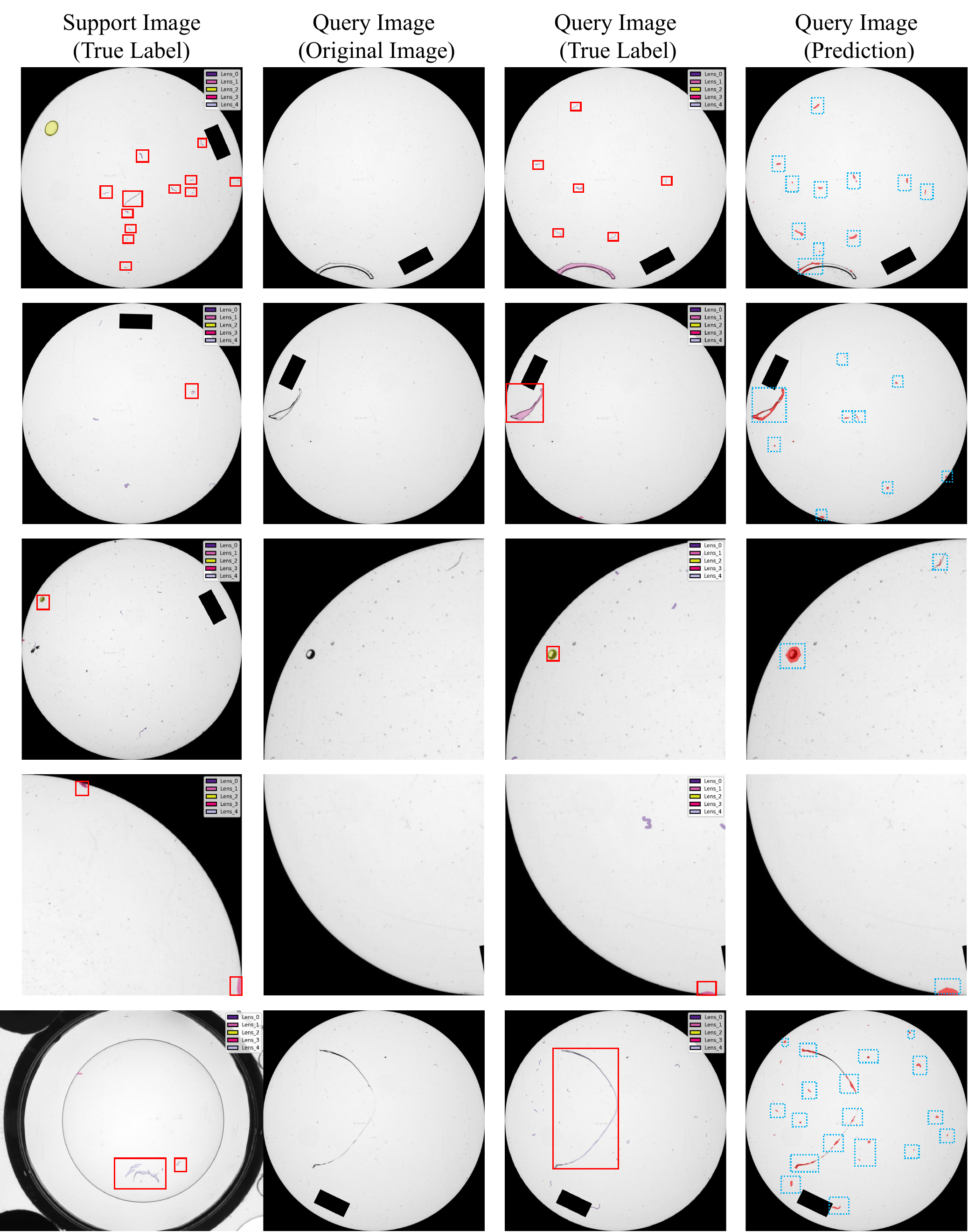}
	\caption{Qualitative results of SOFS under 1-shot setting. Corresponding to Lens\_0, Lens\_1, Lens\_2, Lens\_3 and Lens\_4 from top to bottom. Red solid line box indicates true labels, and blue dashed line box indicates predictions. Due to the small defects, please use the electronic version to enlarge defects for easier viewing.}
	\label{vis: app_fig_6}
\end{figure}

\begin{figure}[t]
	\centering
	\includegraphics[width=\linewidth]{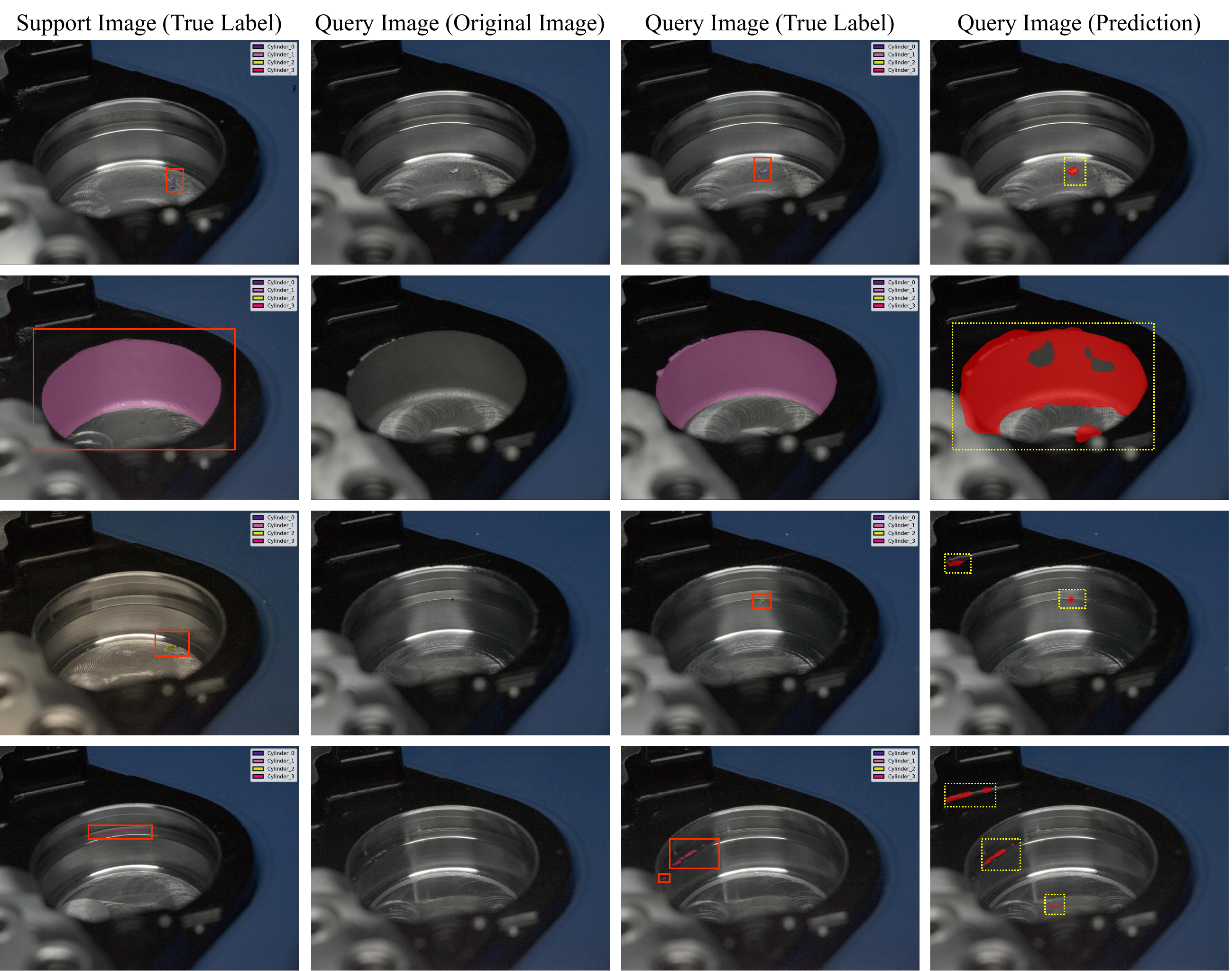}
	\caption{Qualitative results of SOFS under 1-shot setting. Corresponding to Cylinder\_0, Cylinder\_1, Cylinder\_2 and Cylinder\_3 from top to bottom. Red solid line box indicates true labels, and yellow dashed line box indicates predictions. Due to the small defects, please use the electronic version to enlarge defects for easier viewing.}
	\label{vis: app_fig_7}
\end{figure}

We show the visualizations for different defects of a query image conditioned on different support images in Fig. \ref{fig_6} and Fig. \ref{fig_7}. This shows the segmentation results of SOFS for the specified defect. We show the qualitative results of every defect type on VISION V1 in Fig. \ref{vis: app_fig_1}, \ref{vis: app_fig_2}, \ref{vis: app_fig_3}, \ref{vis: app_fig_4}, \ref{vis: app_fig_5}, \ref{vis: app_fig_6}, \ref{vis: app_fig_7}.

\newpage

\end{document}